%% file: main.tex
\documentclass[10pt,twocolumn,letterpaper]{article}

\usepackage[pagenumbers]{iccv} 

\input{preamble}

\definecolor{iccvblue}{rgb}{0.21,0.49,0.74}
\usepackage[pagebackref,breaklinks,colorlinks,allcolors=iccvblue]{hyperref}

\usepackage{xcolor}         
\usepackage{graphicx} 
\usepackage{booktabs} 
\usepackage{array} 
\usepackage{makecell}
\usepackage{colortbl}
\usepackage{multicol}
\usepackage{multirow}

\definecolor{mygray}{gray}{0.2}
\definecolor{lightred}{rgb}{1, 0.5, 0.5}
\definecolor{lightyellow}{rgb}{1, 0.7, 0.2}

\definecolor{Gray}{gray}{0.9}
\definecolor{mygreen}{rgb}{0.0, 0.5, 0.0}
\definecolor{myred}{rgb}{0.8, 0.25, 0.33}
\definecolor{myblue}{rgb}{0.19, 0.55, 0.91}
\definecolor{uclablue}{rgb}{0.15, 0.45, 0.68}
\definecolor{boxgreen}{rgb}{0.02, 0.66, 0.02}
\definecolor{boxred}{rgb}{0.66, 0.1, 0.1}
\definecolor{boxblue}{rgb}{0.01, 0.01, 0.73}
\definecolor{lightgreen}{rgb}{0.745,0.85,0.68} 
\definecolor{lightorange}{rgb}{0.98,0.76,0.65} 

\def\paperID{*****} 
\def\confName{ICCV}
\def\confYear{2025}

\title{Beyond Simple Edits: X-Planner for Complex Instruction-Based Image Editing}

\author{Chun-Hsiao Yeh\textsuperscript{1,3} \quad 
Yilin Wang\textsuperscript{3} \quad 
Nanxuan Zhao\textsuperscript{3} \quad 
Richard Zhang\textsuperscript{3} \quad 
Yuheng Li\textsuperscript{2} \\
Yi Ma\textsuperscript{1,2} \quad 
Krishna Kumar Singh\textsuperscript{3} \quad \\ \\
\textsuperscript{1}UC Berkeley \quad
\textsuperscript{2}HKU \quad
\textsuperscript{3}Adobe
}

\begin{document}

\twocolumn[{
\renewcommand\twocolumn[1][]{#1}%
\maketitle
\vspace{-10mm}
\begin{center}
    \centering
    \includegraphics[width=0.89\linewidth]{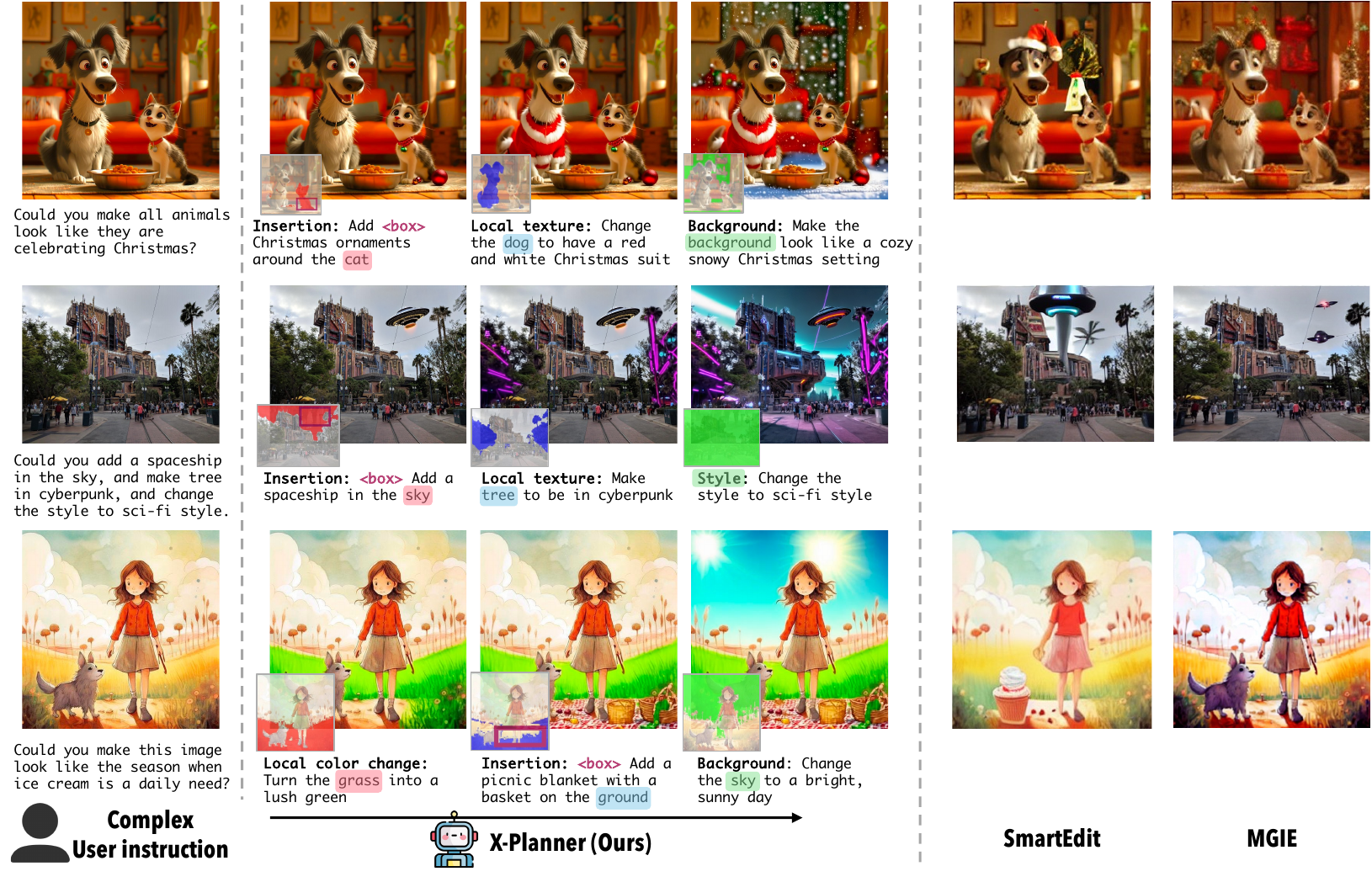}
    \vspace{-5pt}
    \captionof{figure}{\textbf{Left.} Given a source image and complex instruction, our MLLM based \textbf{\emph{X-Planner}} decomposes the complex instruction into simpler sub-instructions (with edit type) along with auto-generated segmentation masks indicating the editing regions (shown in bottom left of each edited image) and  hallucinates additional bounding box of object for the insertion case. We iteratively perform localized editing, by providing \emph{X-Planner}'s editing instruction and region (mask and box) to compatible editing model for each edit type. \textbf{Right.} Recent SmartEdit~\cite{huang2024smartedit} and MGIE~\cite{fu2023guiding} which also use MLLM struggles with complex instruction understanding and identity preservation.}
    \label{fig:teaser}
\end{center}
}]

{\let\thefootnote\relax\footnote{$^*$Work done during CHY's summer internship at Adobe Research.}}

\input{sec/0_abstract}    
\input{sec/1_intro}
\input{sec/2_related}
\input{sec/3_method}

\input{sec/4_exp}

\input{sec/5_conclusion}

{
    \small
    \bibliographystyle{ieeenat_fullname}
    \bibliography{main}
}

\input{supp}

\end{document}

%% file: preamble.tex
\usepackage{xspace}

\usepackage{amsmath,amsfonts,bm}

\usepackage{bbm}
\usepackage{bm}

\usepackage{booktabs}       
\usepackage{amsfonts}       
\usepackage{nicefrac}       
\usepackage{microtype}      
\usepackage{color}
\usepackage{array}
\usepackage{multirow}
\usepackage{dirtytalk}
\usepackage[para,online,flushleft]{threeparttable}
\usepackage{colortbl}

\usepackage[most]{tcolorbox}
\newtcolorbox{prompt}{
    colback=gray!10,
    colframe=gray!40,
    fonttitle=\bfseries,
    coltitle=black,
    colbacktitle=gray!40,
    enhanced,
    drop shadow=black!5!white,
    left=8mm,
    right=8mm,
    top=3mm,
    bottom=3mm,
    boxsep=1mm,
    title=Text Prompt:
    }
\newtcolorbox{gpt_template*}{
    colback=gray!10,
    colframe=gray!40,
    fonttitle=\bfseries,
    coltitle=black,
    colbacktitle=gray!40,
    enhanced,
    drop shadow=black!5!white,
    left=8mm,
    right=8mm,
    top=3mm,
    bottom=3mm,
    boxsep=1mm,
    title=Complex-Simple Instruction Pair Generation Template (Prompting GPT-4o):
    }

\newtcolorbox{gpt_template2*}{
    colback=gray!10,
    colframe=gray!40,
    fonttitle=\bfseries,
    coltitle=black,
    colbacktitle=gray!40,
    enhanced,
    drop shadow=black!5!white,
    left=8mm,
    right=8mm,
    top=3mm,
    bottom=3mm,
    boxsep=1mm,
    title=Complex-Simple Instruction Pair Generation Template (GPT-4o) - Part 2:
    }

\newtcolorbox{fs_query}{
    colback=gray!10,
    colframe=gray!40,
    fonttitle=\bfseries,
    coltitle=black,
    colbacktitle=gray!40,
    enhanced,
    drop shadow=black!5!white,
    left=8mm,
    right=8mm,
    top=3mm,
    bottom=3mm,
    boxsep=1mm,
    title=Few-Shot Queries:
    }
\newtcolorbox{bkg_template}{
    colback=gray!10,
    colframe=gray!40,
    fonttitle=\bfseries,
    coltitle=black,
    colbacktitle=gray!40,
    enhanced,
    drop shadow=black!5!white,
    left=8mm,
    right=8mm,
    top=3mm,
    bottom=3mm,
    boxsep=1mm,
    title=Few-Shot Background Template:
    }
\newtcolorbox{bkg_query}{
    colback=gray!10,
    colframe=gray!40,
    fonttitle=\bfseries,
    coltitle=black,
    colbacktitle=gray!40,
    enhanced,
    drop shadow=black!5!white,
    left=8mm,
    right=8mm,
    top=3mm,
    bottom=3mm,
    boxsep=1mm,
    title=Few-Shot Background Queries:
    }
\newtcolorbox{bkg_prompt}{
    colback=gray!10,
    colframe=gray!40,
    fonttitle=\bfseries,
    coltitle=black,
    colbacktitle=gray!40,
    enhanced,
    drop shadow=black!5!white,
    left=8mm,
    right=8mm,
    top=3mm,
    bottom=3mm,
    boxsep=1mm,
    title=Retrieved Background:
    }

%% file: sec/0_abstract.tex
\begin{abstract}

Recent diffusion-based image editing methods have significantly advanced text-guided tasks but often struggle to interpret complex, indirect instructions. Moreover, current models frequently suffer from poor identity preservation, unintended edits, or rely heavily on manual masks. To address these challenges, we introduce \textbf{X-Planner}, a Multimodal Large Language Model (MLLM)-based planning system that effectively bridges user intent with editing model capabilities. \textbf{X-Planner} employs chain-of-thought reasoning to systematically decompose complex instructions into simpler, clear sub-instructions. For each sub-instruction, \textbf{X-Planner} automatically generates precise edit types and segmentation masks, eliminating manual intervention and ensuring localized, identity-preserving edits. Additionally, we propose a novel automated pipeline for generating large-scale data to train  \textbf{X-Planner} which achieves state-of-the-art results on both existing benchmarks and our newly introduced complex editing benchmark. The project page is available at \url{https://danielchyeh.github.io/x-planner/}.

\end{abstract}

%% file: sec/1_intro.tex
\section{Introduction}
\label{sec:intro}

The field of generative image editing has experienced remarkable advancements in recent years~\cite{nichol2021glide,lugmayr2022repaint,hertz2022prompt,avrahami2022blended,cao2023masactrl,brooks2023instructpix2pix}, driven by the development of diffusion models~\cite{rombach2022high,podell2023sdxl}.  Broadly, these advancements in image editing can be categorized into two categories. The first involves free-form editing, adapts pre-trained diffusion models to edit images based solely on text instructions and a source image~\cite{brooks2023instructpix2pix,zhang2024magicbrush,sheynin2024emu}. These methods often suffer from over-editing, modifying regions beyond those intended by the user. Subsequent research involve controllable image editing that incorporates user-provided control signals—such as semantic segmentation masks~\cite{nichol2021glide,avrahami2022blended,zhao2024ultraedit}, bounding boxes~\cite{chen2024training,li2023gligen,wang2024instancediffusion}, content dragging and blobs~\cite{mou2023dragondiffusion,shi2024dragdiffusion,nie2024compositional}, and image prompts~\cite{ye2023ip,chen2024anydoor} to guide task-specific edits. These signals improve editing control but are time-consuming for users to provide manually.

Another critical challenge of these  models is to robustly interpret and execute complex instructions. Existing methods often struggle with nuanced requirements of these complex instructions, which limits their ability to to perform these edits effectively and provide intuitive, user-friendly interactions {to allow more direct user controls over these edits. Below, we identify several key challenges that highlight limitations of current editing models: \textbf{\textit{(1) Multi-Object Targeting Instructions:}} A single instruction might targets multiple objects within an image which requires editing model to identify each object which needs to be edited according to this single instruction and image content (e.g., first row in Figure~\ref{fig:teaser}). \textbf{\textit{(2)Multi-Task Instructions:}} Instructions with multiple distinct edits within a single prompt (e.g., second row in Figure~\ref{fig:teaser}). \textbf{\textit{(3) Indirect Instruction Interpretation:}} Instructions that contain indirect cues which require deeper understanding and decomposition into multiple steps to achieve accurate results (e.g., last row in Figure~\ref{fig:teaser}).

Recent diffusion-based image editing methods have significantly advanced text-guided tasks, yet still struggle to robustly interpret and execute complex, indirect instructions. Typically, these tasks require extensive manual effort to simplify instructions and provide precise region guidance, limiting scalability and usability. While MLLMs offer promise due to extensive world knowledge, directly applying them often results in misinterpretation and localization errors (as shown in Figure~\ref{fig:teaser}, SmartEdit~\cite{huang2024smartedit} and MGIE~\cite{fu2023guiding} struggle to perform these complex edits as it misunderstand the editing prompt ,e.g., SmartEdit generates ice-cream, instead making it summer scene). One solution is to force these MLLMs to reason about these complex instructions based on image content and apply chain-of-thought reasoning to autonomously break down complex instructions into simpler sub-instructions aligned with image context. In addition, we need these MLLMs to provide region guidance in terms of masks or bounding boxes for these edits. Some existing MLLMs like LISA~\cite{lai2024lisa} and GLaMM~\cite{rasheed2024glamm} provide grounding masks while answering image-related questions but have not been trained for localizing editing instructions.

Recent work, GenArtist~\cite{wang2024genartist}, explored this direction by leveraging closed-source GPT-4~\cite{achiam2023gpt} for instruction decomposition. However, it has several key limitations: (1) It focuses on breaking down long, nested prompts rather than reasoning about complex and indirect instructions, limiting its effectiveness for intricate editing tasks. (2) GenArtist relies on closed-source model and external toolboxes even during inference, restricting adaptability for users and hindering fine-tuning or further research advancements for the community. (3) Most importantly, its dependence on external object detectors~\cite{liu2024grounding} and segmentation model~\cite{kirillov2023segment} results in bounding boxes and masks that are not optimized for different editing types, leading to failures in tasks requiring more than simple object detection. For instance, in object insertion tasks (e.g., "add a cat") where the cat is not originally present in the image, external detectors and segmentation models (e.g., SAM~\cite{kirillov2023segment}) struggle to hallucinate and localize the inserted object, failing to provide effective editing guidance.

To address these challenges, we propose \textbf{\emph{X-Planner}}, a Multimodal Large Language Model (MLLM)-driven planning system that excels in managing complex instruction-based image editing tasks as shown in Figure~\ref{fig:teaser}. Our \emph{X-Planner} operates by breaking down intricate user instructions into structured, simpler sub-instructions, each accompanied by auto-generated edit type and mask corresponding to main editing anchor (e.g., in second row of Figure~\ref{fig:teaser},\textit{“local texture: Make $<$tree$>$ to be in cyberpunk ”} indicates the edit type to be of texture change and main editing anchor is tree for which we generate mask). This decomposition strategy empowers the model to interpret ambiguous requests and apply stepwise complex edits.

Our approach refines spatial control by tailoring masks to each edit type—tight for color/texture changes, coarse for replacements, and full-image for global edits—preventing over-editing and preserving content outside the mask. In the case of insertion edits, just the mask for editing anchor is not sufficient as the inserted object region can be outside the mask. As shown in Figure~\ref{fig:teaser}, the inserted ornaments have to be around the cat which is beyond the mask of cat's body, for which GenArtist~\cite{wang2024genartist} by relying on external detector and segmentor would often fail to handle. \emph{X-Planner} addresses this issue by leveraging the MLLM world knowledge and reasoning about the image content predicts an additional bounding box indicating the region where the object could be inserted. 
\emph{X-Planner} also predicts edit type which enables dynamically selecting the most suitable editing model for each edit type.

\begin{figure*}[!h]
    \centering
    \includegraphics[width=\linewidth]{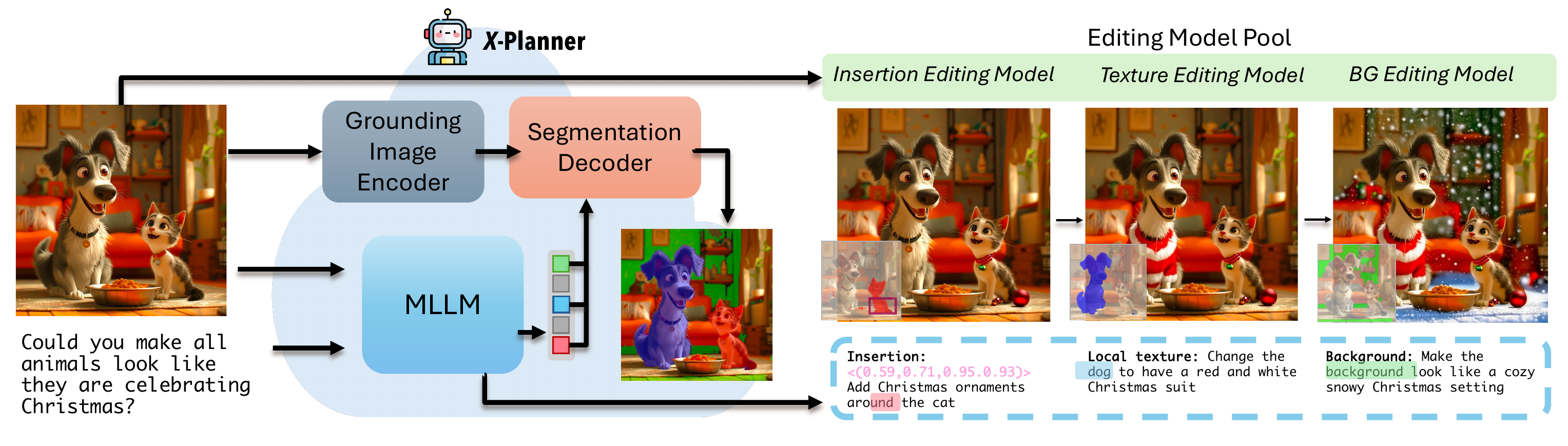}
    \vspace{-15pt}
    \caption{\textbf{Overview of \emph{X-Planner} for Complex Instruction-Based Editing. } Our \emph{X-Planner} comprises two main branches: First, MLLM decomposes the complex instruction into multiple simpler sub-instructions along with editing anchors (e.g., cat, dog, and background) which are given to segmentation decoder to get corresponding masks for each sub-instruction. Also, for the insertion edit, MLLM outputs bounding box coordinates along with edit instruction. By integrating with the editing model pool, we then iteratively apply the most suitable editing model to execute each specific edit task based on \emph{X-Planner} generated sub-instruction along with masks / bounding boxes.}
    
    \label{fig:xplanner-overview}
    \vspace{-10pt}
\end{figure*}

In order to train such planner, we need a large-scale training data mapping complex instruction to simple instructions with edit type, segmentation masks, and bounding boxes for insertion edits. Currently, no such dataset exists, hence we created large-scale \textit{Complex Instruction-based Editing Dataset (COMPIE)}. It comprises of over 260K paired complex-simple instructions along with mask and bounding box annotations for insertion edit. COMPIE is designed with our novel automated data annotation pipeline and stringent quality verification processes, ensuring the dataset is both scalable and highly reliable for evaluating editing capabilities. 
In light of the lack of benchmarks for evaluating complex instruction-driven image editing, we also propose a comprehensive evaluation protocol and a benchmark focusing on complex instructions.  To sum up, our contributions can be summarized as follows:

\begin{itemize}

    \item \textbf{\emph{X-Planner}: Novel end-to-end, self-contained MLLM-based planner to support complex image editing.} 
    
    We introduce \emph{X-Planner}, an MLLM-driven agent that automatically decomposes complex user instructions into simpler tasks, with auto-generated masks and boxes for insertion edit task without relying on external MLLMs and detectors or segmentors during test time.
    
     \item \textbf{A fully automated pipeline for creating large-scale training dataset for complex editing planning}. 
    To support the training of \emph{X-Planner}, we present an automatic large-scale dataset creation pipeline for generating complex-simple instruction pairs, segmentation masks, bounding boxes, and edit types.

    \item \textbf{New complex instruction-based editing benchmark.} 
    We introduce a large-scaled curated benchmark, COMPIE that targets compositional and indirect instructions requiring world knowledge that aims to catalyze on real-world editing tasks beyond existing editing benchmarks. 

\end{itemize}

%% file: sec/2_related.tex
\section{Related Works}
\label{sec:related-work}

\noindent\textbf{Controllable Generative Image Editing.}
Text-to-image diffusion models have recently demonstrated remarkable performance in generating high-quality images from textual descriptions~\cite{rombach2022high,podell2023sdxl}. Building on this success, pre-trained diffusion models have been adapted for image editing tasks guided by editing text input. Some of these are training free methods~\cite{meng2021sdedit,hertz2022prompt,parmar2023zero,cao2023masactrl,hertz2024style,wu2024turboedit} like SD-Edit~\cite{meng2021sdedit}, Prompt-to-Prompt~\cite{hertz2022prompt}, and Pix2PixZero~\cite{parmar2023zero} which perform text-based image editing by injecting noise into the image and then guiding the diffusion process to align with the editing text. Another line of works are training based methods~\cite{brooks2023instructpix2pix,zhang2024magicbrush,fu2023guiding,ge2024seed} like InstructPix2Pix~\cite{brooks2023instructpix2pix}, MagicBrush~\cite{zhang2024magicbrush} which are more robust than training-free methods and relies on creating paired data of original and edited image to fine-tune the text-to-image diffusion models. While these methods improve upon earlier approaches, they often suffer from over-editing, affecting regions unrelated to the user’s instructions. Subsequent methods~\cite{nichol2021glide,avrahami2022blended,zhao2024ultraedit,chen2024training,li2023gligen,wang2024instancediffusion,mou2023dragondiffusion,shi2024dragdiffusion,nie2024compositional,ye2023ip,chen2024anydoor} have improved controllable image editing by incorporating additional control signals, such as segmentation masks, bounding boxes, dragging, blobs, and image prompts. However, these methods often require users to provide control guidance manually, which can be laborious and limits usability. Also, they are generally limited to simple, direct instructions and struggle with more complex ones. Our \emph{X-Planner} overcomes these challenges by automatically decomposing user instructions into actionable sub-instructions, generating the segmentation masks and bounding boxes as control guidance.

\noindent\textbf{MLLM-based Image Editing.}
Multimodal Large Language Models (MLLMs) leverage the strengths of the large language models (LLMs)~\cite{chowdhery2023palm,touvron2023llama,dubey2024llama} while incorporating visual data~\cite{liu2024visual, zhu2023minigpt}, enabling more sophisticated multimodal understanding and generation. Recent efforts have extended MLLMs to the domain of image editing~\cite{huang2024smartedit,fu2023guiding}. E.g., MGIE~\cite{fu2023guiding} uses MLLMs to make editing instructions more expressive and use that to guide editing models such as InstructPix2Pix to have better image editing ability. Also, a very recent concurrent work, GenArtist~\cite{wang2024genartist} uses an external closed-source GPT-4~\cite{achiam2023gpt} as agent that decomposes long editing tasks with multiple simple instructions nested together (similar to multi-task instructions mentioned in the introduction) and uses external object localization tools. In contrast, our proposed \emph{X-Planner} is an MLLM agent which handles actual indirect complex instructions and generates editing type specific masks which are beyond simple object detection (e.g. for insertion it hallucinates object location based on image contnet). Moreover, \emph{X-Planner} operates independently at inference, without relying on large, closed-source models like GPT-4 for planning.

%% file: sec/3_method.tex
\section{Method}
\label{sec:method}

In this section, we introduce \emph{X-Planner} (Figure~\ref{fig:xplanner-overview}), a method specifically designed to break down complex instructions into simpler image editing tasks. Leveraging an MLLM trained on our proposed large-scale dataset tailored for instruction decomposition, \emph{X-Planner} autonomously generates control inputs—such as segmentation masks and bounding boxes—for each sub-instruction to facilitate precise and instruction-based image editing.

\subsection{X-Planner: A Complex Editing Task Agent}

We first break down this task of complex-instruction based image editing planning into key sub-problems of (1) complex instruction decomposition, and (2) control guidance input generation. Figure~\ref{fig:xplanner-overview} presents an overview of our \emph{X-Planner} pipeline, which first decompose complex instructions into multiple simpler instructions along with corresponding control guidance inputs (mask for all edits and bounding box for insertion edit). We then conduct the iterative editing by assigning the suitable editing model for each edit task. We draw inspiration from the recently introduced GLaMM~\cite{rasheed2024glamm}, which is an MLLM that utilizes an LLaVA-like architecture~\cite{liu2024visual} with a segmentation mask decoder. This model constructs image-level captions with specific phrases linked to corresponding segmentation masks. For example, given an image with a cat and a dog, GLaMM can respond to \textit{`can you describe this image?'} with \textit{`there are a $<$cat$>$ and a $<$dog$>$'}, anchored to unique segmentation masks for each animal. So, GLaMM would use vision-language understanding of MLLM to generate the caption along with anchored phrases (like cat and dog) and then segmentation mask decoder takes the input image feature and these anchored phrases to segment them. 

In order to adapt the GLaMM for our case, we would like it to take a source image and a complex instruction, break it down to simpler instructions with edit type and anchored editing object/region (e.g. cat, dog, and background in Figure~\ref{fig:xplanner-overview}). Thus, segmentation decoder can take these anchored editing regions to get corresponding masks. For the insertion case, we also want to predict the location of the object to be inserted but we cannot use the segmentation decoder as it can only segment the object visible in the image. Hence, we would use MLLM  which has world-knowledge to predict bounding box based on the input image and insertion instruction (e.g. \textit{[insertion]$<$ 0.59,0.71,0.95,0.93$>$ Add Christmas ornaments around the cat} in Figure~\ref{fig:xplanner-overview}).

However, we realize that GLaMM does not work well for complex instruction-based planning tasks. Specifically, (1) it struggles with complex instruction decomposition, as it primarily learns from visual grounding samples rather than the instruction interpretation, and (2) it is limited in generating control guidance inputs, such as task-specific masks, especially for insertion tasks where it fails to hallucinate unseen objects. These two main problems remain unsolved in the current GLaMM, thus motivating the need for a large-scale, complex image editing instruction planning dataset to train MLLM and segmentation decoder of GLaMM model to generalize for this specific task.

\begin{figure}[!t]
    \centering
    \includegraphics[width=1\columnwidth]{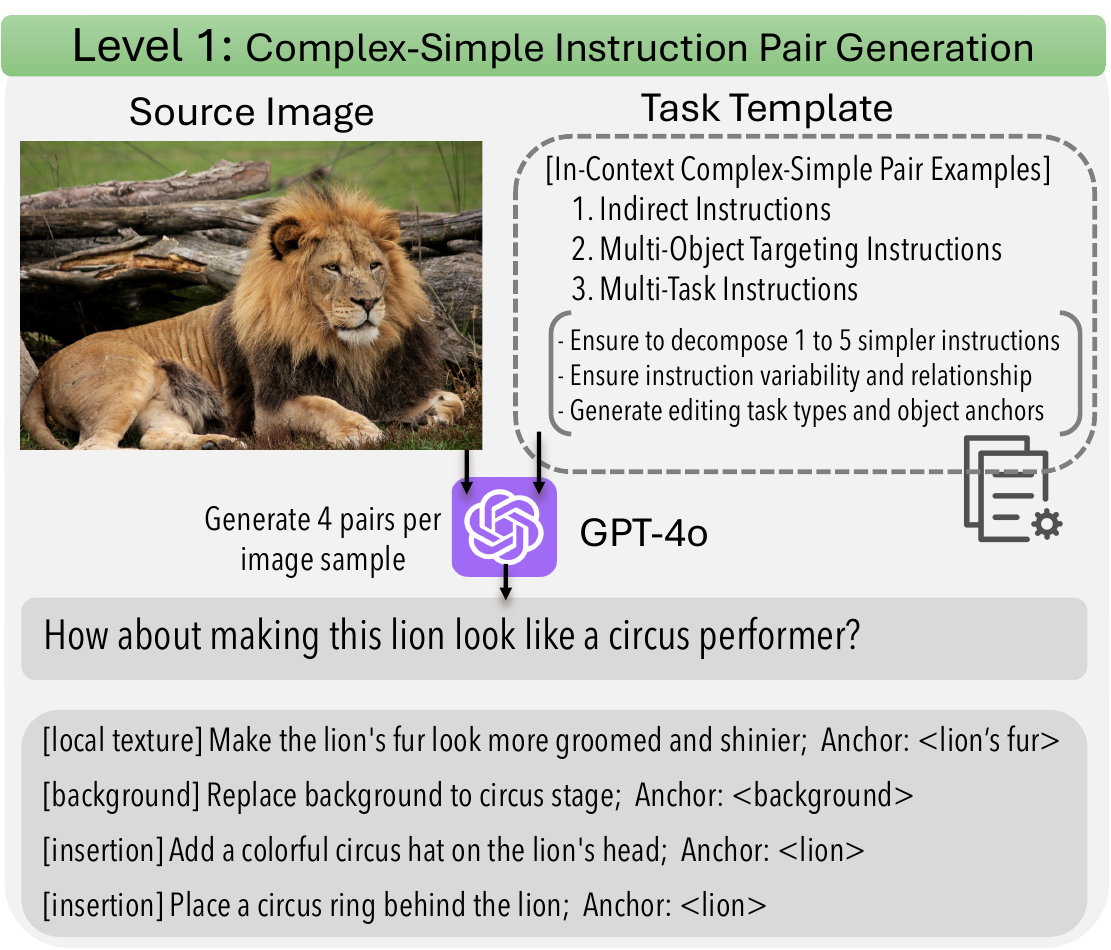}
    \vspace{-15pt}
    \caption{\textbf{Level 1: Complex-Simple Instruction Pair Generation.} Using our structured template, we prompt GPT-4o to generate complex instructions—including indirect, multi-object, and multi-task instructions (as defined in Section 1)—along with their corresponding simpler instructions, object anchors, and edit types.
    } 
    \label{fig:data-instruction}
    \vspace{-10pt}
\end{figure}

\subsection{Automated Data Annotation Pipeline}

We present our novel automated annotation pipeline developed to construct the Complex Instruction-Based Editing Dataset (COMPIE), a comprehensive and diverse dataset for complex instruction-driven editing planner. 

\begin{figure}[!t]
    \centering
    \includegraphics[width=0.96\columnwidth]{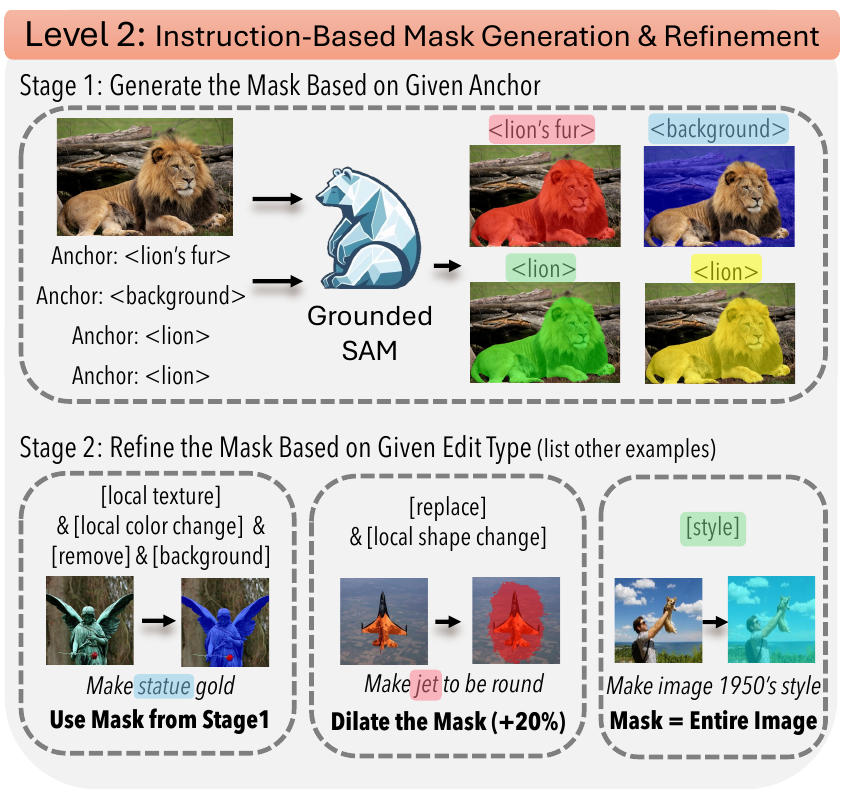}
    \vspace{-5pt}
    \caption{\textbf{Level 2: Instruction-Based Mask Generation and Refinement.} In Stage 1, we use the source image and anchor text with Grounded SAM to generate a fine-grained mask for the specified object. In Stage 2, we refine this mask by applying varies strategies based on the edit type provided in Level 1 (Figure~\ref{fig:data-instruction}).
    } 
    \label{fig:data-mask}
\vspace{-10pt}
\end{figure}

\begin{figure}[!t]
    \centering
    \includegraphics[width=1\columnwidth]{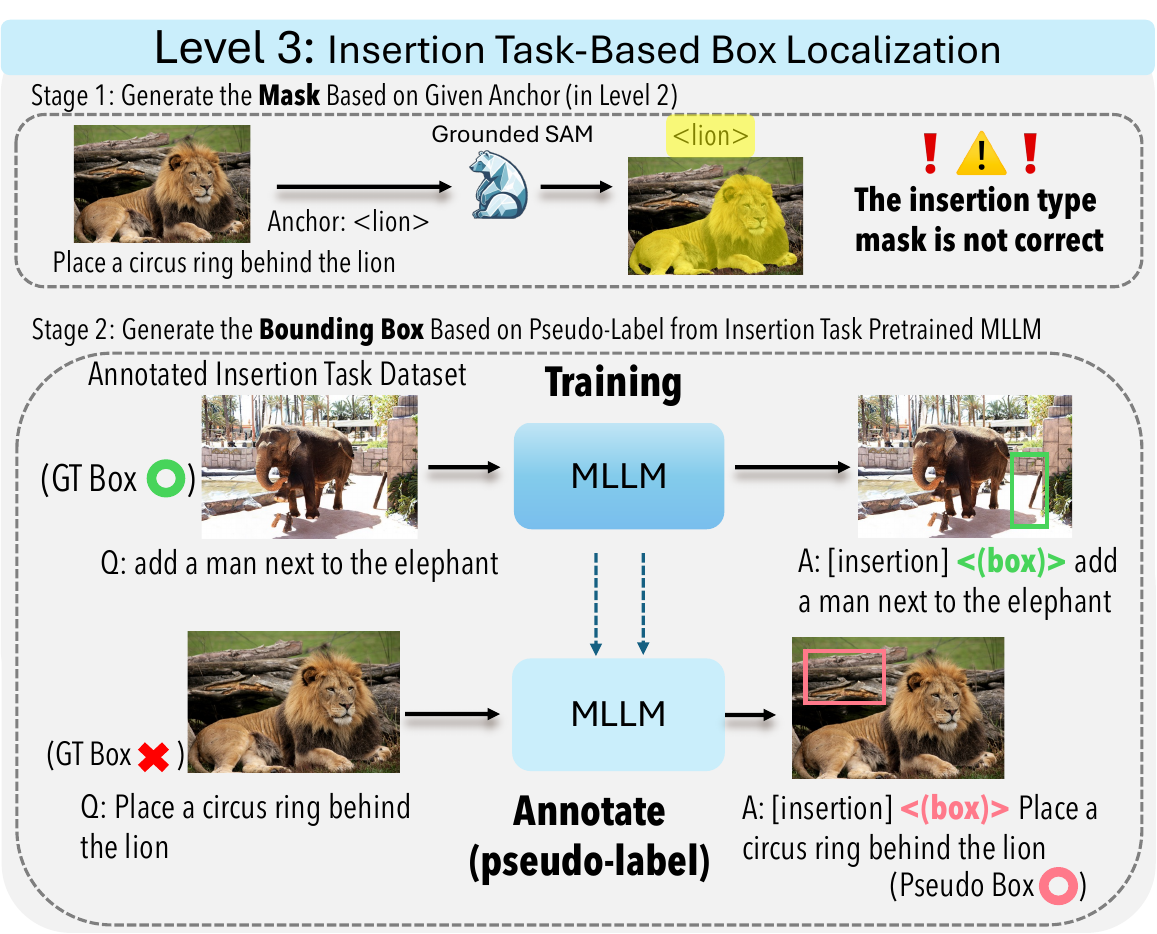}
    \vspace{-15pt}
    \caption{\textbf{Level 3: Insertion Task-Based Mask \& Box Localization.} For insertion task, Grounded SAM struggles to segment objects not present in the source image. We pre-train an MLLM on a bounding box-annotated dataset~\cite{tudosiu2024mulan}, enabling it to pseudo-annotate our data with bounding box for insertion edits.
    } 
    \label{fig:data-box}
    \vspace{-10pt}
\end{figure}

The pipeline comprises three distinct levels. At Level 1 (Figure~\ref{fig:data-instruction}), we generate structured complex and decomposed simple instruction (with editing anchors) pairs using GPT-4o~\cite{achiam2023gpt} for an input image. Level 2 (Figure~\ref{fig:data-mask}) employs Grounded-SAM~\cite{ren2024grounded} to produce initial segmentation masks anchored to each instruction, providing a base for spatial control, and these masks are further refined according to edit type. Level 3 (Figure~\ref{fig:data-box}) focuses on insertion-based instructions to generate precise bounding boxes indicating location where object can be inserted.

\noindent\textbf{Level 1: Complex-Simple Instruction Pair Generation.} 
To address the challenge of limited instruction diversity, we generate complex editing instructions by leveraging MLLM creativity and human oversight. We draw on diverse data sources—including SEED-X~\cite{ge2024seed}, UltraEdit~\cite{zhao2024ultraedit}, and InstructPix2Pix~\cite{brooks2023instructpix2pix} datasets—spanning synthetic to real images and varying quality levels. First, we manually design in-context examples that capture key instruction types: indirect, multi-object, and multi-task instructions (as defined in Section 1). We ensure that each complex instruction decomposes into 1 to 5 simpler sub-instructions with meaningful and coherent correspondence, and that each sub-instruction specifies an editing task type and editing anchor which corresponds to edited object/region. Each simple edit will have one anchor except replace edit which will have two anchors corresponding to objects before and after replace edit. Using GPT-4o~\cite{achiam2023gpt}, we then generate four complex-simple instruction pairs per image by prompting it with both the source image and our designed task template (see Figure~\ref{fig:data-instruction}). Also, in our training data we mix simple-simple instruction pairs to ensure \emph{X-Planner} learns to not modify or breakdown simple instructions. Our approach also works using the open-sourced models like Pixtral-Large~\cite{agrawal2024pixtral}. Please see Supp. for more details.

\noindent\textbf{Level 2: Instruction Mask Generation and Refinement.} 
\emph{X-Planner} aims to provide editing models with precisely defined target regions for modification. At Level 2 (Stage 1), we generate segmentation masks based on the edited object anchor identified in each decomposed instruction from Level 1 (e.g., \textit{“add a circus ring behind the lion; Anchor: $<$lion$>$”}). Using the source image and anchor text, we employ Grounded SAM~\cite{ren2024grounded} to produce a fine-grained mask for the specified object.

In Stage 2, we refine the mask based on the edit type (see Figure~\ref{fig:data-instruction}). For local texture, color change, and background edits, we use the mask generated in Stage 1 directly. For shape changes, we make masks larger to accommodate object reshaping by dilating the mask by 20\%. In replace tasks (e.g., \textit{“replace cat with dog”}), we take the union of both masks (e.g. cat and dog) based on the anchor objects from pre- and post-edit images if available (e.g., InstructPix2Pix dataset). For datasets like SEED-X, where only the pre-edit image is available, we dilate the mask by 20\% to account for the replace edit. For style changes or instructions requiring global transformation (e.g., \textit{“make image 1950's style”}), we select the entire image as the editing mask.

\noindent\textbf{Level 3: Insertion Task-Based Box Localization.} 
In insertion-type edits (e.g., \textit{“add a circus ring behind the lion”} as shown in Figure~\ref{fig:data-instruction}), Grounded SAM, which relies on object anchors, encounters difficulties when segmenting objects that are not present in the source image (e.g., \textit{“the circus ring”}). Direct segmentation of the intended placement region (e.g., \textit{“lion”}) often leads to imprecise masks, particularly when there is ambiguity between the instruction and the mask (e.g., \textit{“add a circus ring behind the lion”}), causing errors by segmenting the existing object (lion) rather than the intended insertion area (behind the lion). This discrepancy can lead to degraded edit quality.

To address these limitations, in Level 3 (Stage 2), we fine-tune the MLLM component of GLaMM~\cite{rasheed2024glamm} on an annotated dataset with ground truth (GT) bounding boxes for insertion locations. Specifically, we leverage the MULAN dataset~\cite{tudosiu2024mulan}, which includes background images with and without foreground objects, allowing us to use images without the object as input and predict the bounding box of the insertion target. Through training, the MLLM learns to generate bounding box recommendations for novel insertion instructions during inference. For unannotated insertion instructions, the fine-tuned MLLM produces pseudo-labels by predicting bounding boxes based on the given instruction, enabling precise edit placements. In our experiments (Section~\ref{sec:exp-box}), we further demonstrate \emph{X-Planner}’s capability to generate consistent bounding box predictions with repeated instruction to show plausible variations in location.

%% file: sec/4_exp.tex
\begin{figure*}[!h]
    \centering
    \includegraphics[width=\linewidth]{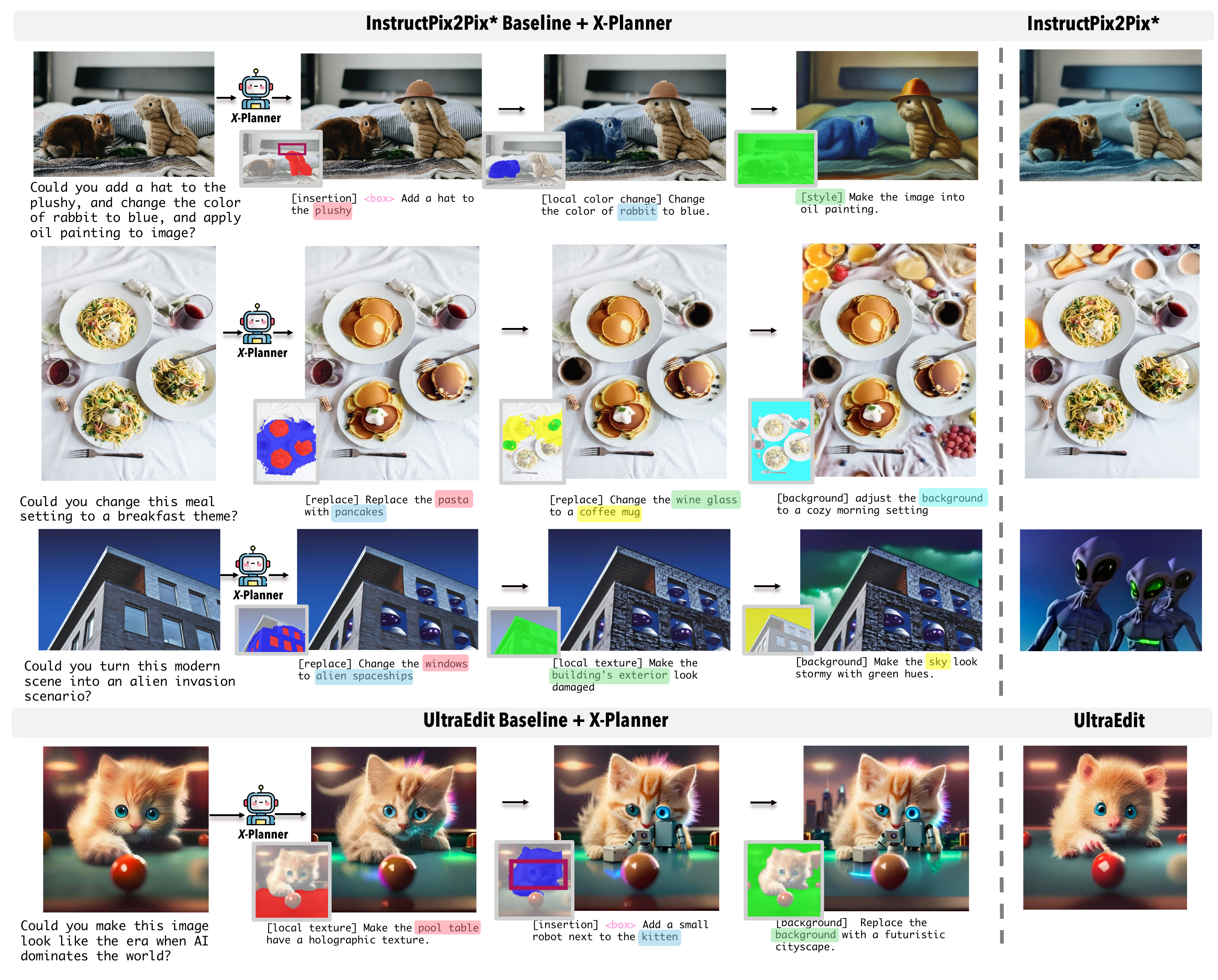}
    \vspace{-15pt}
    \caption{\textbf{Qualitative Comparison for Complex Instruction-Based Editing Benchmark.} Integrating \emph{X-Planner} with editing methods, InstructPix2Pix* and UltraEdit, brings drastic boosts in preserving object identities with \emph{X-Planner} generated masks and boxes (display in bottom-left of each image). \emph{X-Planner}'s decomposition of complex instructions also enhances alignment with various complex instruction inputs. \emph{X-Planner} provides a distinct advantage over baselines that only use the source image and complex instruction without masks.} 
    \label{fig:xplanner-compare}
    \vspace{-10pt}
\end{figure*}

\section{Experiments}
\label{sec:exp}

Our experiments evaluate the quality of \emph{X-Planner} in complex instruction understanding and editing localization. First, we test performance on simple instruction settings using the established MagicBrush benchmark~\cite{zhang2024magicbrush}. Second, we evaluate complex instruction settings by comparing performance with and without plugging in \emph{X-Planner} to baselines on our proposed COMPIE benchmark.

\noindent\textbf{Settings.}
\emph{X-Planner} uses GLaMM~\cite{rasheed2024glamm} as base model, built on Vicuna-7B~\cite{zheng2023judging}. For training, over 260K complex-simple instruction pairs COMPIE is used. See supp. for more details.

\noindent\textbf{Baselines.}
For the MagicBrush benchmark, which focuses on simple instruction settings, we benchmark against methods tailored for straightforward edits~\cite{brooks2023instructpix2pix,zhang2024hive,zhang2024magicbrush,meng2021sdedit,nichol2021glide,avrahami2022blended,zhao2024ultraedit}. We explore variations that integrate components of \emph{X-Planner} into the UltraEdit~\cite{zhao2024ultraedit} baseline, which is current state-of-the-art and can also take editing mask as input to assess the impact of \emph{X-Planner} in simpler editing scenarios. For the COMPIE benchmark, which addresses complex instruction settings, we select UltraEdit~\cite{zhao2024ultraedit} and InstructPix2Pix* as primary baselines to show the effectiveness of our \emph{X-Planner}.  InstructPix2Pix* is an improved version of InstructPix2Pix~\cite{brooks2023instructpix2pix} using our internal dataset and also utilizes mask as input conditioning. We use this model to show generalizability of our \emph{X-Planner} due to lack of public models with mask conditioning.  We evaluate these methods in two variants: with \emph{X-Planner} integration and without it. Also, we compare with MGIE~\cite{fu2023guiding} and SmartEdit~\cite{huang2024smartedit}, which use MLLM to improve the editing performance.

\noindent\textbf{Benchmark and Metrics.}
For the MagicBrush benchmark~\cite{zhang2024magicbrush}, we use its evaluation setup, and for our edits measure L1, L2 distance, CLIP-I, and DINO similarity with ground-truth. For the COMPIE benchmark, we adopt the evaluation protocol from EmuEdit~\cite{sheynin2024emu}, comparing edited images against both the source image and target captions. Consistent with EmuEdit, we use L1, CLIP image similarity ($CLIP_{im}$), and DINO similarity to measure how well the edited image retains the content of the original image. And use CLIP text-image similarity ($CLIP_{out}$) to measure alignment of editing instruction and edited image.   Given that $CLIP_{out}$ can struggle to capture the nuances of complex instructions, we additionally employ InternVL2-Llama3-76B~\cite{chen2024internvl2}, a powerful MLLM, to evaluate the alignment between the editing instruction and edited image ($MLLM_{ti}$). We use InternVL2 for fair evaluation instead of GPT-4o as GPT-4o was used to create our training data. For completeness, we also use InternVL2 to measure the similarity of input image and edited image ($MLLM_{im}$). Please find Supp. for more details.

\begin{table}[!h]
\small
\centering
\caption{\textbf{Quantitative Comparison on the MagicBrush Test Set.} We report results for both single-turn and multi-turn settings. In comparison to UltraEdit using human labeled masks, we evaluate using \emph{X-Planner} generated masks and bounding boxes as control inputs. \textit{For Bag of Models, we utilize PowerPaint for removal tasks, InstructDiff for style changes, and UltraEdit for other edit types.}}

\resizebox{0.48\textwidth}{!}{
\begin{tabular}{lccccc}
\toprule
\multicolumn{6}{c}{\textbf{Single-Turn}} \\ 
\midrule
\textbf{Methods} & \textbf{Guidance Control Input} & L1$\downarrow$ & L2$\downarrow$ & CLIP-I$\uparrow$ & DINO$\uparrow$ \\ 
\midrule

SD-SDEdit & No & 0.1014 & 0.0278 & 0.8526 & 0.7726 \\

Null Text Inversion & No & 0.0749 & 0.0197 & 0.8827 & 0.8206 \\ \cmidrule{1-6} 

GLIDE & Human Labeled Mask & 3.4973 & 115.8347 & 0.9487 & 0.9206 \\
Blended Diffusion & Human Labeled Mask & 3.5631 & 119.2813 & 0.9291 & 0.8644 \\ \cmidrule{1-6}
HIVE             & No & 0.1092 & 0.0380 & 0.8519 & 0.7500 \\
InstructPix2Pix (IP2P) & No & 0.1141 & 0.0371 & 0.8512 & 0.7437 \\
IP2P w/ MagicBrush & No & 0.0625 & 0.0203 & 0.9332 & 0.8987 \\
\cellcolor{gray!30}UltraEdit & \cellcolor{gray!30}No   & \cellcolor{gray!30}0.0614 & \cellcolor{gray!30}0.0181 & \cellcolor{gray!30}0.9197 & \cellcolor{gray!30}0.8804 \\
\cellcolor{gray!30}UltraEdit & \cellcolor{gray!30}Human Labeled Mask     & \cellcolor{gray!30}0.0575 & \cellcolor{gray!30}0.0172 & \cellcolor{gray!30}0.9307 & \cellcolor{gray!30}\textbf{0.8982} \\ 
\cellcolor{blue!5}\emph{X-Planner} + UltraEdit & \cellcolor{blue!5}\emph{X-Planner}'s Mask & \cellcolor{blue!5}0.0528 & \cellcolor{blue!5}0.0171 & \cellcolor{blue!5}0.9281 & \cellcolor{blue!5}0.8900 \\
\cellcolor{blue!5}\emph{X-Planner} + UltraEdit & \cellcolor{blue!5}\emph{X-Planner}'s Mask + Box & \cellcolor{blue!5}0.0513 & \cellcolor{blue!5}\textbf{0.0168} & \cellcolor{blue!5}0.9312 & \cellcolor{blue!5}0.8959 \\
\cellcolor{blue!5}\emph{X-Planner} + Bag of Models & \cellcolor{blue!5}\emph{X-Planner}'s Mask + Box & \cellcolor{blue!5}\textbf{0.0511} & \cellcolor{blue!5}0.0172 & \cellcolor{blue!5}\textbf{0.9331} & \cellcolor{blue!5}0.8970 \\

\toprule

\multicolumn{6}{c}{\textbf{Multi-Turn}} \\ 
\midrule
\textbf{Methods} & \textbf{Guidance Control Input} & L1$\downarrow$ & L2$\downarrow$ & CLIP-I$\uparrow$ & DINO$\uparrow$ \\ 
\midrule

SD-SDEdit & No & 0.1616 & 0.0602 & 0.7933 & 0.6212 \\
Null Text Inversion & No & 0.1057 & 0.0335 & 0.8468 & 0.7529 \\ \cmidrule{1-6} 
GLIDE & Human Labeled Mask & 11.7487 & 1079.5997 & 0.9094 & 0.8494 \\
Blended Diffusion & Human Labeled Mask & 14.5439 & 1510.2271 & 0.8782 & 0.7690 \\ \cmidrule{1-6} 
HIVE             & No & 0.1521 & 0.0557 & 0.8004 & 0.6463 \\
InstructPix2Pix (IP2P) & No & 0.1345 & 0.0460 & 0.8304 & 0.7018 \\
IP2P w/ MagicBrush & No & 0.0964 & 0.0353 & 0.8924 & 0.8273 \\
\cellcolor{gray!30}UltraEdit, eval w/o region  & \cellcolor{gray!30}No  & \cellcolor{gray!30}0.0780 & \cellcolor{gray!30}0.0246 & \cellcolor{gray!30}0.8954 & \cellcolor{gray!30}0.8322 \\
\cellcolor{gray!30}UltraEdit, eval w/ region   & \cellcolor{gray!30}Human Labeled Mask  & \cellcolor{gray!30}0.0745 & \cellcolor{gray!30}0.0236 & \cellcolor{gray!30}0.9045 & \cellcolor{gray!30}0.8505 \\
\cellcolor{blue!5}\emph{X-Planner} + UltraEdit & \cellcolor{blue!5}\emph{X-Planner}'s Mask & \cellcolor{blue!5}0.0679 & \cellcolor{blue!5}0.0227 & \cellcolor{blue!5}0.9025 & \cellcolor{blue!5}0.8423 \\
\cellcolor{blue!5}\emph{X-Planner} + UltraEdit & \cellcolor{blue!5}\emph{X-Planner}'s Mask + Box & \cellcolor{blue!5}0.0668 & \cellcolor{blue!5}0.0226 & \cellcolor{blue!5}0.9047 & \cellcolor{blue!5}0.8475 \\
\cellcolor{blue!5}\emph{X-Planner} + Bag of Models & \cellcolor{blue!5}\emph{X-Planner}'s Mask + Box & \cellcolor{blue!5}\textbf{0.0665} & \cellcolor{blue!5}\textbf{0.0223} & \cellcolor{blue!5}\textbf{0.9079} & \cellcolor{blue!5}\textbf{0.8508} \\
\bottomrule
\end{tabular}
}
\label{tab:magicbrush_result}
\end{table}

\subsection{MagicBrush Results (Simple Instructions)}
Table~\ref{tab:magicbrush_result} shows quantitative results on the MagicBrush benchmark. Key observations: (1) \emph{X-Planner} enhances UltraEdit by providing masks and bounding boxes for localized edits, improving performance, especially for insertion tasks. \emph{X-Planner}'s mask is able to match the human labeled mask, and even outperform it sometimes as shown in the result. (2) \emph{X-Planner} is model-agnostic, integrating seamlessly with multiple models (e.g., PowerPaint~\cite{zhuang2023task} for removal, InstructDiff~\cite{geng2024instructdiffusion} for style changes, and UltraEdit for other edits), leveraging their strengths to boost overall performance. Please find Supp. for more quantitative results.

\subsection{COMPIE-Eval Results (Complex Instructions)}

\noindent\textbf{Qualitative Results.} 
In Figure~\ref{fig:xplanner-compare}, we show complex instruction editing results for InstructPix2Pix* and UltraEdit with and without \emph{X-Planner}. We can see that without \emph{X-Planner} editing methods are not able to understand the user intentions from complex instructions. For example, in the last row of Figure~\ref{fig:xplanner-compare}, the intention of the instruction was to make image look futuristic but just UltraEdit fails to understand that whereas our \emph{X-Planner} is able to convert complex instruction into meaningful sub-instructions. Even for the cases where editing method understands the meaning of instructions, they struggle with identity preservation as they cannot leverage the editing masks and bounding boxes of the planner (e.g. third row for InstructPix2Pix* in Figure~\ref{fig:xplanner-compare}).

\begin{table*}[!t]
    
    \centering
    \caption{\textbf{Quantitative Comparison on the COMPIE Benchmark.} \emph{X-Planner} significantly improves the editing performance of UltraEdit and InstructPix2Pix* by decomposing complex instructions and providing control guidance inputs (e.g., masks). To overcome the limitations of $CLIP_{out}$ in handling complex instructions, we utilize an MLLM-based evaluation metric to better reflect \emph{X-Planner}'s capabilities.}

    \vspace{-5pt}

    \resizebox{0.8\textwidth}{!}{
    \begin{tabular}{lccccc|cc}
    \toprule
     \textbf{Methods} & \textbf{Guidance Control Input} & L1$\downarrow$ & CLIP$_{im}$ $\uparrow$ & CLIP$_{out}$$\uparrow$ & DINO$\uparrow$ & MLLM$_{ti}$ $\uparrow$ & MLLM$_{im}$ $\uparrow$ \\ 
    \midrule
                SmartEdit~\cite{huang2024smartedit} & No & 0.2764 & 0.7713 & 0.2512 & 0.6044 & 0.6511 & 0.5347 \\  
                  MGIE~\cite{fu2023guiding} & No & 0.2988 & 0.7692 & 0.2498 & 0.5981 & 0.6408 & 0.5288 \\  
                  \midrule
                  \cellcolor{gray!30}UltraEdit & \cellcolor{gray!30}No & \cellcolor{gray!30}0.1292 & \cellcolor{gray!30}0.7688 & \cellcolor{gray!30}\textbf{0.2698} & \cellcolor{gray!30}0.6387 & \cellcolor{gray!30}0.6652 & \cellcolor{gray!30}0.5523 \\

                  \cellcolor{blue!5}GenArtist~\cite{wang2024genartist} + UltraEdit & \cellcolor{blue!5}GenArtist's Mask + Decomposed Instruction & \cellcolor{blue!5}0.1253 & \cellcolor{blue!5}0.7767 & \cellcolor{blue!5}0.2621 & \cellcolor{blue!5}0.6435 & \cellcolor{blue!5}0.6894 & \cellcolor{blue!5}0.5593 \\

                  \cellcolor{blue!5}\emph{X-Planner} + UltraEdit & \cellcolor{blue!5}\emph{X-Planner}'s Decomposed Instruction & \cellcolor{blue!5}0.1253 & \cellcolor{blue!5}0.7767 & \cellcolor{blue!5}0.2621 & \cellcolor{blue!5}0.6435 & \cellcolor{blue!5}0.6894 & \cellcolor{blue!5}0.5593 \\ 
                  
                  \cellcolor{blue!5}\emph{X-Planner} + UltraEdit & \cellcolor{blue!5}\emph{X-Planner}'s Mask + Decomposed Instruction & \cellcolor{blue!5}\textbf{0.1188} & \cellcolor{blue!5}\textbf{0.7875} & \cellcolor{blue!5}0.2569 & \cellcolor{blue!5}\textbf{0.6599} & \cellcolor{blue!5}\textbf{0.7061} & \cellcolor{blue!5}\textbf{0.5744} \\

                  \midrule
                  \cellcolor{gray!30}InstructPix2Pix* & \cellcolor{gray!30}No & \cellcolor{gray!30}0.1517 & \cellcolor{gray!30}0.8020 & \cellcolor{gray!30}\textbf{0.2666} & \cellcolor{gray!30}0.6988 & \cellcolor{gray!30}0.6727 & \cellcolor{gray!30}0.6160 \\

                  \cellcolor{blue!5}GenArtist~\cite{wang2024genartist} + InstructPix2Pix* & \cellcolor{blue!5}GenArtist's Mask + Decomposed Instruction & \cellcolor{blue!5}0.1458 & \cellcolor{blue!5}0.8143 & \cellcolor{blue!5}0.2641 & \cellcolor{blue!5}0.7114 & \cellcolor{blue!5}0.7072 & \cellcolor{blue!5}0.6277 \\

                  \cellcolor{blue!5}\emph{X-Planner} + InstructPix2Pix* & \cellcolor{blue!5}\emph{X-Planner}'s Decomposed Instruction & \cellcolor{blue!5}0.1458 & \cellcolor{blue!5}0.8143 & \cellcolor{blue!5}0.2641 & \cellcolor{blue!5}0.7114 & \cellcolor{blue!5}0.7072 & \cellcolor{blue!5}0.6277 \\

                  \cellcolor{blue!5}\emph{X-Planner} + InstructPix2Pix* & \cellcolor{blue!5}\emph{X-Planner}'s Mask + Decomposed Instruction & \cellcolor{blue!5}\textbf{0.1320} & \cellcolor{blue!5}0.8285 & \cellcolor{blue!5}0.2591 & \cellcolor{blue!5}0.7068 & \cellcolor{blue!5}\textbf{0.7408} & \cellcolor{blue!5}\textbf{0.6454} \\
    
    \bottomrule
    \end{tabular}
    }
\vspace{-5pt}
\label{tab:COMPIE_result}
\end{table*}

\noindent\textbf{Quantitative Results.}
We create a high quality and diverse test benchmark, COMPIE-Eval, focusing on complex editing by collecting data from a different sources, LAION-high-aesthetics~\cite{schuhmann2022laion}, and Unsplash-2K~\cite{kim2021noise}. We then use GPT-4o to generate complex instruction for the test image and apply a post-verification stage, in which crowd workers filter examples with irrelevant instructions. The COMPIE contains 550 images, with complex instructions with variations mentioned in the introduction. See supp. for details.

In Table~\ref{tab:COMPIE_result}, we show that \emph{X-Planner} enhances editing performance by simplifying complex instruction to decomposed simple instructions suitable for editing models and generating masks helps it further for improved identity preservation. It improves UltraEdit and InstructPix2Pix* across most metrics, except $CLIP_{out}$, which struggles with edited image and complex instruction alignment (e.g., focusing on ice cream instead of the summer scene in Figure~\ref{fig:teaser} last row). To address this, we use InternVL2, an MLLM-based text-image alignment metric ($MLLM_{ti}$), which better understands complex instructions, showing significant improvements with \emph{X-Planner}. MGIE~\cite{fu2023guiding} and SmartEdit~\cite{huang2024smartedit} despite using MLLM gives inferior results. Please note that InternVL2 can also be used for verifying the quality of the edited results and choose the best one.

\begin{figure}[!h]
    \centering
    \includegraphics[width=0.9\columnwidth]{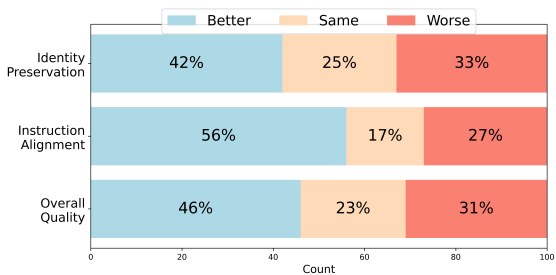}
     \vspace{-10pt}
    \caption{\textbf{User Study on COMPIE Benchmark.} We compare against InstructPix2Pix* and UltraEdit. “Better” means the generated images by using our \emph{X-Planner} is preferred and vice versa. 
    } 
    \label{fig:user-study}
\end{figure}

\begin{table}[t!]
    \small 
    \scriptsize

    \caption{\textbf{Segmentation Mask Comparison on PIE Benchmark}. We compare on the setting of instruction-to-segmentation mask, \emph{X-Planner} consistently outperforms baseline methods.}
    \vspace{-5pt}
    \centering
    \renewcommand{\arraystretch}{0.9} 
    \setlength{\tabcolsep}{4pt} 
    
    \resizebox{0.7\columnwidth}{!}{
    \begin{tabular}{l|ccc}
    \toprule
    Method & IoU$\uparrow$ & Precision$\uparrow$ & Recall$\uparrow$ \\
    \cmidrule(lr){1-4} 

        Random 10\% Mask  & 0.09 & 0.49 & 0.12 \\
        \midrule
        GLaMM-Base  & 0.14 & 0.66 & 0.15 \\
        GLaMM-RefSeg  & 0.28 & 0.69 & 0.32 \\
        Llama3+GLaMM  & 0.44 & 0.73 & 0.53 \\
        \cellcolor{blue!5}\emph{X-Planner} (Ours)  & \cellcolor{blue!5}\textbf{0.67} & \cellcolor{blue!5}\textbf{0.79} & \cellcolor{blue!5}\textbf{0.81} \\

    \bottomrule
    \end{tabular}
    }
    \label{tab:x-planner-seg}
\end{table}

\noindent\textbf{User Study.} We conducted a user study with 100 random samples from 550 images in the COMPIE benchmark to compare results of two baselines: InstructPix2Pix* and UltraEdit with and without \emph{X-Planner}. Participants rated each image on (1) identity preservation, (2) instruction alignment, and (3) overall quality, and choose the preferred image or rate them equal. In Figure~\ref{fig:user-study}, we show average results for both benchmarks and observe the users prefer results with \emph{X-Planner} for all criteria (better means \emph{X-Planner} preferred).

\begin{figure}[!h]
    \centering
    \includegraphics[width=1\columnwidth]{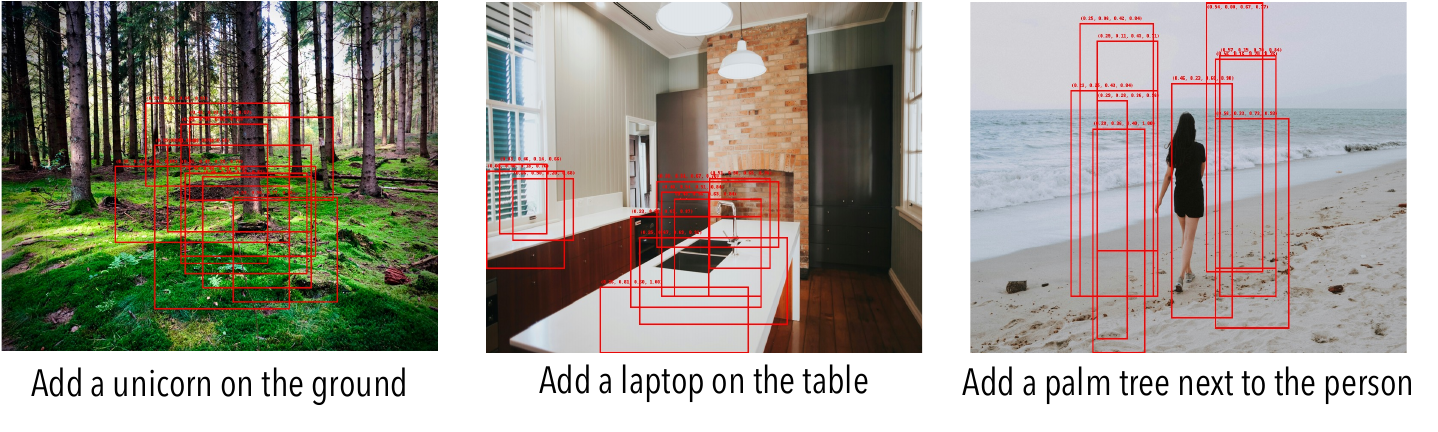}
    \vspace{-10pt}
    \caption{\textbf{Visualize Consistent Bounding Box with Repeated Runs.} We show \emph{X-Planner} can generate consistent bounding boxes with repeated runs to yield plausible location variations.
    } 
    \label{fig:box-visuals}
\end{figure}

\subsection{GLaMM Comparison and BBox Localization}
\label{sec:exp-box}

In Table~\ref{tab:x-planner-seg}, we report \emph{X-Planner}’s segmentation mask generation performance on the PIE benchmark~\cite{ju2024pnp}, comparing it with GLaMM~\cite{rasheed2024glamm} variations (baseline model and a version fine-tuned on RefSeg dataset to have better grounding) and a baseline that leverages Llama3~\cite{dubey2024llama} for object anchoring followed by GLaMM for mask generation. Based on the results, we can see \emph{X-Planner} consistently surpasses other instruction-to-segmentation methods, underscoring its effectiveness for mask generation in instruction-based tasks.

We visualize \emph{X-Planner}'s bounding box localization for insertion edits in Figure~\ref{fig:box-visuals} with 10 repeated runs of the same instruction. Predicted box locations (laptop on table) and shapes (vertical for palm tree) appear plausible. Detailed quantitative analysis and ablations are in the Supp.

%% file: sec/5_conclusion.tex
\section{Conclusion}
In this paper, we introduced \emph{X-Planner}, an MLLM-based planning system that breaks down complex instructions into simpler tasks with editing masks and bounding boxes for insertion edits. We also proposed a novel data generation pipeline to train this planner. Our evaluation highlights \emph{X-Planner}'s potential to enhance existing editing models, encouraging further exploration of MLLM-based planners as complementary tools for complex editing tasks.

%% file: supp.tex
\clearpage
\setcounter{page}{1}

\newcommand{\methodname}{\textit{X-Planner}\xspace}

\section*{Overview} 
In this supplementary, we first provide additional details about our training dataset and proposed evaluation benchmark in Section~\ref{sec:supp-compie-summary} and \emph{X-Planner}'s implementation details in Section~\ref{sec:supp-implementation}. Next, we show some more quantitative results for \emph{X-Planner}'s bounding box guidance ability in Section~\ref{sec:supp-exp-box}. Then, We demonstrate our planner trained with open-sourced model and show various qualitative comparisons with baseline methods to show the effectiveness of our planner in Section~\ref{sec:supp-emu},~\ref{sec:supp-error},~\ref{sec:supp-pixtral-data},~\ref{sec:supp-pixtral-result},~\ref{sec:supp-qual-result}, and~\ref{sec:supp-non-rigid}.

\begin{figure*}[!t]
    \centering
    \includegraphics[width=\linewidth]{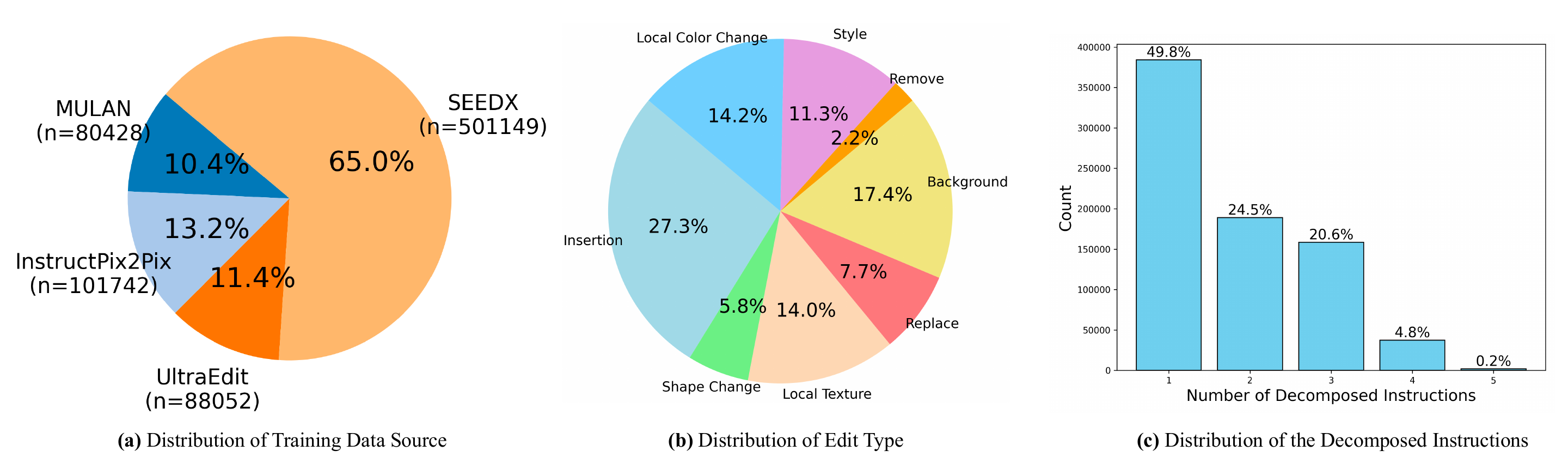}
    \caption{\textbf{\emph{X-Planner}'s Training Dataset Summary (Generated from GPT-4o).} This figure provides a distribution summary of our \emph{X-Planner}'s training dataset, including (a) the data sources, (b) the distribution of edit types for decomposed instructions, and (c) the number of decomposed instructions. The dataset demonstrates significant diversity and comprises large-scaled number of pairs. Note that $n$ in (a) indicates the number of data samples.}

    \label{fig:supp-compie-summary}
\end{figure*}

\begin{figure*}[!t]
    \centering
    \includegraphics[width=0.9\linewidth]{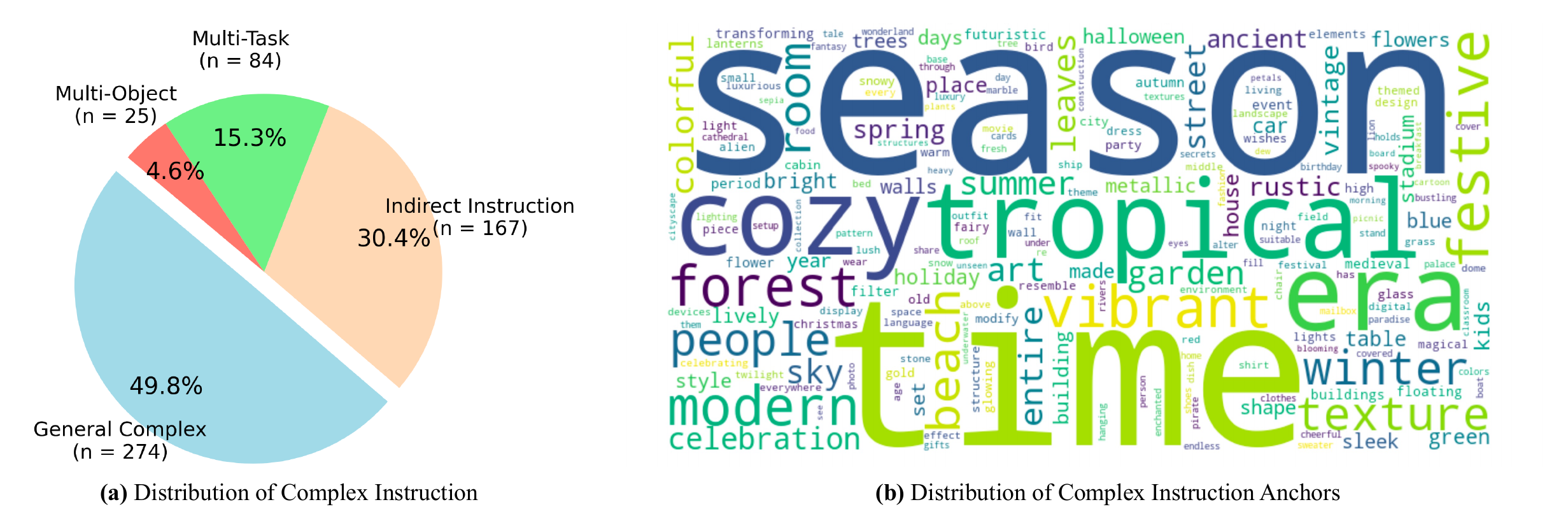}
    \caption{\textbf{COMPIE Benchmark Summary.} This figure summarizes the proposed COMPIE benchmark, which consists of 550 samples spanning various types of complex instructions shown in (a), including (1) general complex instructions, (2) indirect instructions, (3) multi-object instructions, and (4) multi-task instructions. Additionally in (b), we present the word count distribution of complex instruction anchor descriptions, highlighting the diversity of the dataset. Note that $n$ in (a) indicates the number of data samples.}

    \label{fig:supp-compie-benchmark}
\end{figure*}

\section{COMPIE Dataset \& Benchmark Summary}
\label{sec:supp-compie-summary}

\noindent\textbf{\emph{X-Planner}'s Training Dataset Statistics (COMPIE).}
We explore the details of our proposed COMPIE, a large-scale and high-quality dataset specifically designed to address complex instruction-based image editing. COMPIE contains over 260K complex-to-simple instruction pairs, focusing on complex editing tasks unlike previous works (e.g., MagicBrush) that predominantly focus on simple instructions. Figure~\ref{fig:supp-compie-summary} provides a comprehensive summary of the dataset, including (a) data sources such as SEEDX~\cite{ge2024seed}, UltraEdit~\cite{zhao2024ultraedit}, MULAN~\cite{tudosiu2024mulan}, and InstructPix2Pix~\cite{brooks2023instructpix2pix}, (b) the distribution of edit types across 8 categories for decomposed instructions, and (c) the breakdown of decomposed instructions ranging from 1 to 5 per complex instruction.

\noindent\textbf{COMPIE Validation Benchmark Statistics.}
In this section, we present a detailed summary of the COMPIE validation benchmark, highlighting its diversity and focus on complex instruction-based image editing. COMPIE benchmark comprises 550 samples from LAION high-aesthetic dataset~\cite{schuhmann2022laion} and Unsplash 2K~\cite{kim2021noise}, providing a rich variety of real-world images to enhance generalization across different domains. The COMPIE benchmark is categorized into four distinct types of complex instructions, including (1) general complex instructions (50\%). General complex instructions typically require multiple editing steps to fulfill the directive. For instance, instructions such as \textit{`Make the image look like a winter wonderland'} or \textit{`Transform the room to appear colorful and lively'} necessitate edits across multiple regions or objects to achieve the desired outcome comprehensively, (2) indirect instructions (30\%), (3) multi-object instructions (15\%), and (4) multi-task instructions (5\%), as shown in Figure~\ref{fig:supp-compie-benchmark} (a). This diverse distribution ensures comprehensive coverage of various complexities encountered in real-world editing tasks. Figure~\ref{fig:supp-compie-benchmark} (b) illustrates the word count distribution of complex instruction anchor descriptions, underscoring the instruction diversity in the dataset. The inclusion of diverse anchor words enriches the dataset's ability to evaluate the interpretation ability of editing models. These insights showcase COMPIE as a robust benchmark designed to advance the development of image editing models which understand and execute complex instructions.

\section{Implementation Details}
\label{sec:supp-implementation}

\noindent\textbf{\methodname's Setup.}
Our \emph{X-Planner} leverages GLaMM~\cite{rasheed2024glamm} as the base model, built on Vicuna-7B~\cite{zheng2023judging}. The approach incorporates key components inspired by GLaMM~\cite{rasheed2024glamm}, particularly the design of the region encoder, grounding image encoder, and pixel decoder. Adhering to the training protocol of GLaMM, we keep the global image encoder and grounding image encoder frozen, while fully fine-tuning the region encoder and pixel decoder. For the language model, we employ LoRA fine-tuning with a scaling factor $\alpha = 8$ over 10 epochs.

\noindent\textbf{MLLM Evaluation Metric Setup.}
In the COMPIE benchmark evaluation, we follow the Emu Edit~\cite{sheynin2024emu} protocol for metrics and employ InternVL2-Llama3-76B~\cite{chen2024internvl2} as our $MLLM$ metric to evaluate instruction-to-edited-image alignment and similarity between the original and edited images. Specifically, we adapt the DreamBench++~\cite{peng2024dreambench++} template for image-to-image alignment by ranking alignment scores from very poor to excellent (0 to 4) based on shape, color, and texture criteria. For instruction-to-image alignment, we ensure that text prompts are complex instructions requiring transformation. For example, the prompt \textit{`Could you make this image look like the season when ice cream is a daily essential?'} is first interpreted as creating a summer scene rather than directly associating the prompt with ice cream. We normalize the maximum score to 4 and calculate the performance percentage based on the average score, presenting it as the $MLLM$ metric performance.

\noindent\textbf{\methodname's Training Dataset Setup.}
For training dataset distribution, we ensure that \emph{X-Planner} retains the ability to handle simple instructions. In the InstructPix2Pix~\cite{brooks2023instructpix2pix} dataset, 40\% of training pairs are simple-to-simple, meaning the decomposition directly maps simple instructions to simple outputs. Similarly, for the MULAN~\cite{tudosiu2024mulan} dataset, which focuses on insertion-type edits, we treat the dataset pairs as simple-to-simple for insertion-specific training. For fine-tuning \emph{X-Planner}, we integrate datasets used in the original GLaMM~\cite{rasheed2024glamm} model, including Semantic\_Segm, RefCoco\_GCG, PSG\_GCG, Flickr\_GCG, and GranDf\_GCG, in combination with our COMPIE training dataset including InstructPix2Pix\_GCG, UltraEdit\_GCG, SEEDX\_GCG, and MULAN\_GCG. The data source ratio for training our \emph{X-Planner} is [1, 3, 3, 3, 1, 3, 3, 9, 9, 9].

\section{\emph{X-Planner}'s Bounding Box Localization}
\label{sec:supp-exp-box}

In Table~\ref{tab:supp-x-planner-box}, we present results on the MULAN validation benchmark~\cite{tudosiu2024mulan}, demonstrating the localization effectiveness of our \emph{X-Planner}. Key observations include: (1) \emph{X-Planner}’s bounding box localization significantly improves with pseudo-labeling, where we annotate training data with bounding box generated by an MLLM pre-trained on insertion tasks (as described in section 3.2 -- Level 3 in the main paper). Mask Only baseline where we rely on segmentation decoder to predict mask of the location where object would be inserted is not sufficient. E.g., if we want to insert hat on person, then just having mask for the person is not sufficient as the hat would be beyond the person mask, hence we need to take the advantage of our MLLM to predict the bounding box on top of the head. (2) Enlarging small bounding boxes ($<$ 5\% of the image size) in the training set further enhances box prediction accuracy, addressing challenges posed by very small bounding boxes that make insertion tasks difficult for diffusion-based editing models. 

Finally, just predicting bounding box gives good localization capability but sometimes it may not cover full-extent of the object and miss some part which might need to be edited when the object is inserted. So, for better coverage of editing location, we also try combining both bounding box and mask which gives slightly better localization than just using bounding box. Also, as editing location guidance, most of the editing methods like UtlraEdit are more robust to bigger editing location compared to smaller one, hence adding mask along with bounding box is a better strategy as it will give larger coverage of the editing area. E.g. in the last row of Figure~\ref{fig:xplanner-compare}, for inserting robot on table, the mask gives the coverage of table where robot would be inserted and bounding box provides additional coverage on top of the table as the inserted robot would be sitting on the table.

\section{Quantitative Comparison on Emu Edit} 
\label{sec:supp-emu}

We also compare the effectiveness of our \emph{X-Planner} on Emu Edit test set~\cite{sheynin2024emu} which is similar to MagicBrush~\cite{zhang2024magicbrush} test set and focuses on simpler instructions. We apply our \emph{X-Planner} with the UltraEdit~\cite{zhao2024ultraedit} model and we can see in Table~\ref{tab:emu_result} that \emph{X-Planner} improves the performance on all the identity preserving metrics ($L1$, $CLIP_{im}$, and $DINO$) due to better instruction localization through the predicted mask and box which would be given as control guidance to UltraEdit. Also, the instruction following performance (measured by $CLIP_{out}$) is similar to the UltraEdit as most of the instructions are simple and do not require \emph{X-Planner} to simplify it further. Apart from UltraEdit results, we also show results of other baseline editing methods like InstructPix2Pix~\cite{brooks2023instructpix2pix}, MagicBrush~\cite{zhang2024magicbrush}, EmuEdit~\cite{sheynin2024emu}, and OmniGen~\cite{xiao2024omnigen} at the top of Table~\ref{tab:emu_result} as reference.

\noindent\textbf{Note:} Emu Edit~\cite{sheynin2024emu} and UltraEdit~\cite{zhao2024ultraedit} highlight that the MagicBrush benchmark introduces biases favoring models trained on its dataset, leading to inflated performance, as the numbers \textcolor{red}{\textbf{highlighted in red}} for $CLIP_{im}$ and $DINO$ metrics in Table~\ref{tab:emu_result}. This overfitting undermines the general editing capabilities of these models on other datasets.

\section{Multi-Step Editing Error Propagation} 
\label{sec:supp-error}

Our \emph{X-Planner} is less prone to errors since it decomposes complex instructions into simpler, model-friendly steps. As stated in the main paper, we can further enhance reliability by introducing a closed-loop verification mechanism using strong MLLMs (e.g., InternVL2.5-38B~\cite{chen2024internvl2} and GPT-4o~\cite{achiam2023gpt}) to evaluate each intermediate result. Specifically, after each editing step, we use the verifier to assign a score from 0--4 that reflects how well the generated image aligns with the current instruction---similar in spirit to the $MLLM_{ti}$ metric introduced in our evaluation. If the score falls below a threshold (e.g., 3), we automatically re-generate the step using a different random seed to recover from potential hallucinations, misalignment.

This mechanism is critical for catching early-stage failures that would otherwise propagate through subsequent steps. We allow a configurable number of retries (e.g., max=1 or 4), striking a balance between quality and efficiency. As shown in Table~\ref{tab:verify_result}, this approach improves instruction-image alignment and identity preservation compared to baselines without verification.

\begin{figure*}[!t]
    \centering
    \includegraphics[width=\linewidth]{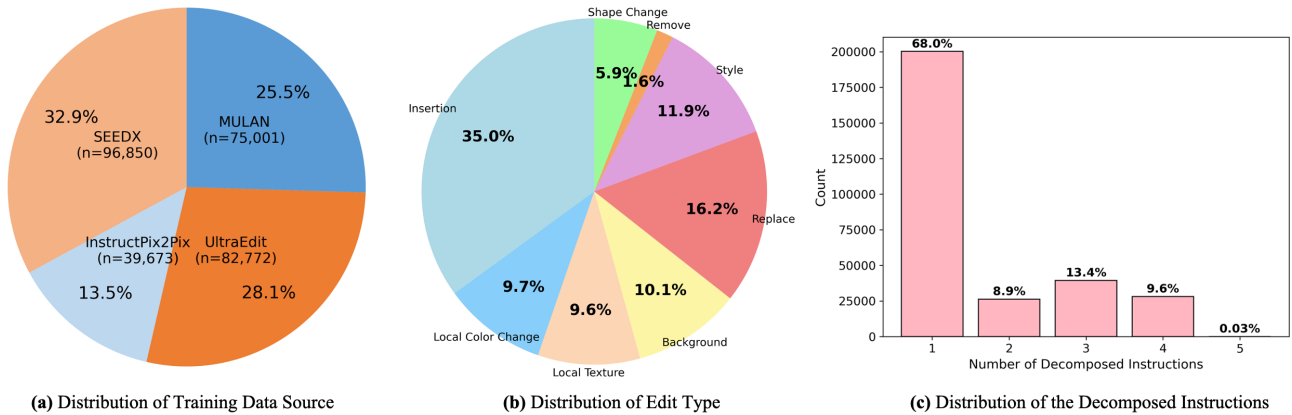}

    \caption{\textbf{\emph{X-Planner}'s Training Dataset Summary (Generated from Pixtral-Large).} This figure illustrates key statistics of our automatically constructed dataset used to train \emph{X-Planner} using \textbf{an open-sourced model}, comprising around 300K instruction-image pairs. 
    \textbf{(a)} \emph{Source Composition:} The data is aggregated from four datasets--SEEDX (32.9\%), UltraEdit (28.1\%), MULAN (25.5\%), and InstructPix2Pix (13.5\%)—with $n$ indicating sample count. 
    \textbf{(b)} \emph{Edit Type Distribution:} Our dataset covers diverse editing intents, including insertion (35.0\%), replace (16.2\%), style (11.9\%), background edits (10.1\%), local texture (9.6\%), local color change (9.7\%), shape change (5.9\%), and remove (1.6\%). This diverse mix supports robust generalization across edit semantics. 
    \textbf{(c)} \emph{Instruction Decomposition Complexity:} While the majority (68.0\%) of prompts require only a single edit, a substantial portion involve multi-step reasoning: 2-step (8.9\%), 3-step (13.4\%), 4-step (9.6\%), and even 5-step (0.03\%). This highlights the need for a sequential planner like \emph{X-Planner} to handle compositional and complex instructions effectively.}

    \label{fig:supp-compie-summary-pixtral}
\end{figure*}

\section{Generate Training Data from Open-Sourced Model, Pixtral-Large}
\label{sec:supp-pixtral-data}

To ensure reproducibility and accessibility of our pipeline, we also build a secondary version of the training dataset using an open-sourced MLLM, Pixtral-Large~\cite{agrawal2024pixtral}, a 124B-parameter multimodal model built upon Mistral Large v2, offering competitive performance on a range of VQA and multimodal reasoning benchmarks. Notably, it surpasses the closed-source GPT-4o~\cite{achiam2023gpt} and Gemini 1.5~\cite{team2024gemini} models on several standard evaluation sets, making it a strong candidate for high-quality instruction generation while remaining fully accessible to the research community.

We deploy Pixtral-Large~\cite{agrawal2024pixtral} using 8 NVIDIA A100 80GB GPUs to enable inference across diverse image-to-instruction generation tasks. The instruction generation process mirrors the methodology used with GPT-4o to ensure a fair comparison: As shown in Figure~\ref{fig:supp-compie-summary-pixtral}, we use the same image datasets, including InstructPix2Pix~\cite{brooks2023instructpix2pix}, UltraEdit-100K~\cite{zhao2024ultraedit}, and SEEDX~\cite{ge2024seed}. The dataset includes around 300K image and instruction pairs. For SEEDX, we generate both complex and simplified instruction pairs from the same image while for InstructPix2Pix, only complex instructions are generated, as simplified ones are already annotated.

\section{Train \emph{X-Planner} with Generated Data from Open-Sourced Model, Pixtral-Large}
\label{sec:supp-pixtral-result}
To ensure a fair comparison, we maintain identical training settings, including model architecture, learning rate, batch size, and number of epochs, as described in Section~\ref{sec:supp-implementation} which were used for GPT-4o version of the \emph{X-Planner}. This allows us to isolate the effect of training signal quality from the underlying data generator. We evaluate the Pixtral-Large-trained \emph{X-Planner} across two benchmarks: 1) MagicBrush test set, and 2) COMPIE benchmark to compare with the GPT-4o version of the \emph{X-Planner}.

Table~\ref{tab:magicbrush_pixtral_result} (MagicBrush Test Set): On both single-turn and multi-turn settings, the Pixtral-trained \emph{X-Planner} sightly outperforms the GPT-4o-trained version. Notably, in the multi-turn evaluation with UltraEdit + Bag of Models, the the Pixtral-Large version of \emph{X-Planner} achieves the highest across all settings. This suggests that Pixtral-generated instruction data have competitive performance in complex editing tasks over multiple rounds.

Table~\ref{tab:COMPIE_pixtral_result} (COMPIE Benchmark): We can see \emph{X-Planner} trained from Pixtral-Large~\cite{agrawal2024pixtral} generated data achieves comparable or slightly improved performance compared to its GPT-4o counterpart across most metrics. For example, using \textit{X-Planner + UltraEdit combined with decomposed instructions with mask control}, Pixtral-Large verson of \emph{X-Planner} yields MLLM$_{ti}$ = 0.7102 and MLLM$_{im}$ = 0.5765, which closely matches GPT-4o’s 0.7061 and 0.5744, respectively. When paired with InstructPix2Pix*, the Pixtral-trained \emph{X-Planner} achieves the best overall MLLM$_{ti}$ and MLLM$_{im}$ score of 0.7431 and 0.6488, indicating stronger consistency in understanding editing outputs. These results demonstrate that training \emph{X-Planner} using data generated from the open-sourced MLLM demonstrating our approach is generalizable and not restricted to the close-sourced GPT-4o.

\section{Additional Qualitative Results}
\label{sec:supp-qual-result}
\begin{figure*}[!h]
    \centering
    \includegraphics[width=1\linewidth]{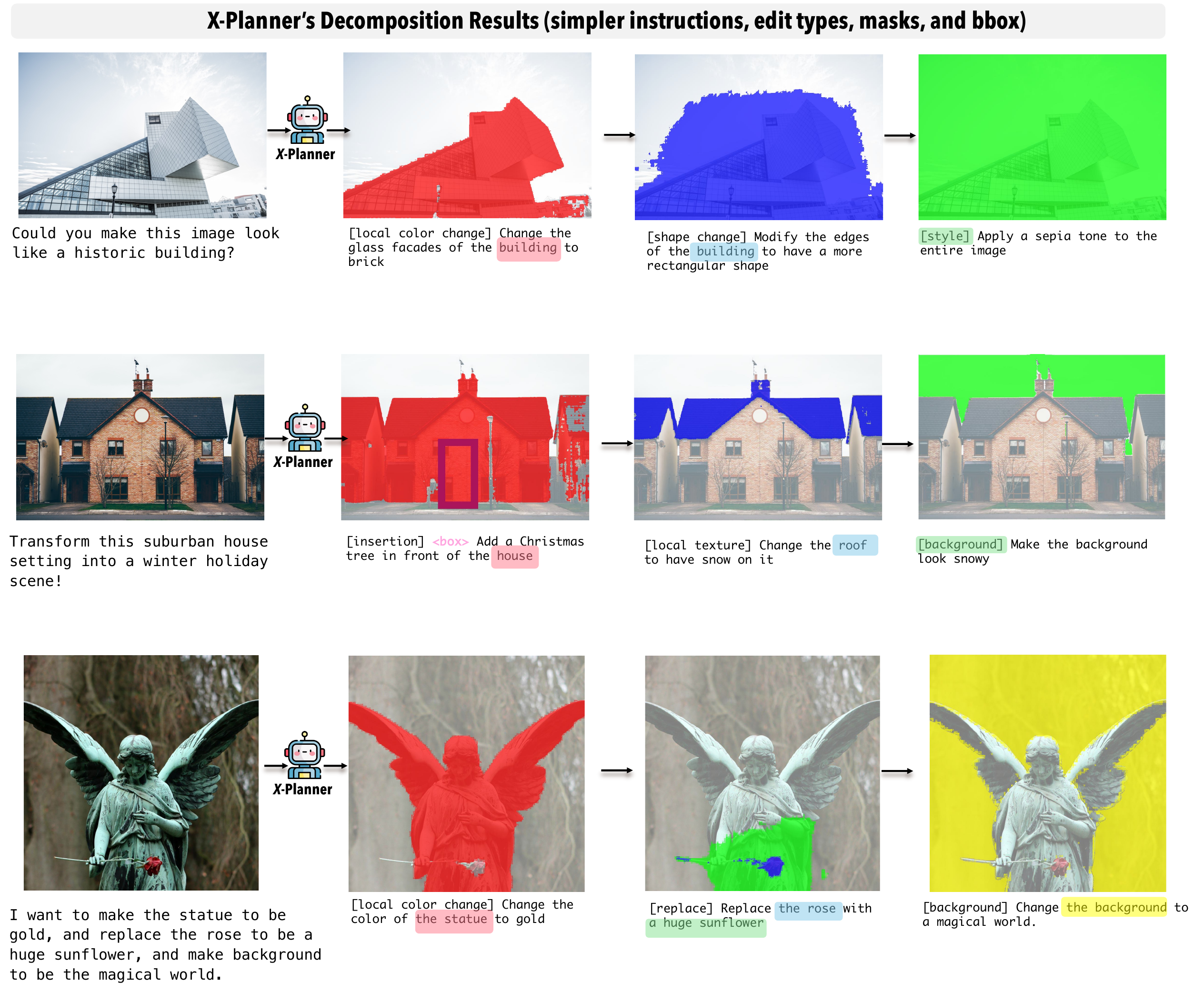}
    \vspace{5pt}
    \caption{\textbf{\emph{X-Planner}'s Decomposition Outputs (Simplified Instructions, Edit Types, Masks, and Bounding Boxes).} This figure illustrates examples of \emph{X-Planner}'s decomposition results, showcasing its ability to handle diverse edit types. The segmentation masks generated are tailored to the specific edit type. For instance, in Row 3, the [replace] edit features both before and after masks which are combined into a single mask for editing guidance, with appropriate dilation applied to ensure accurate representation. Also, in Row 1, the [shape change] edit provides more dilated mask by giving more rooms for potential shape modification.}

    \label{fig:supp-xplanner-outcome}
\end{figure*}

In Figure~\ref{fig:supp-xplanner-outcome}, we provide a detailed look at \emph{X-Planner}'s decomposition capabilities, emphasizing its ability to effectively manage a wide range of edit types by generating precise and context-aware segmentation masks. Each example demonstrates how \emph{X-Planner} tailors its outputs to the specific requirements of the edit type. For instance, in Row 3, the [replace] edit features a sophisticated approach where both the "before" and "after" masks are generated, combined into a single guidance mask. This unified mask, further enhanced with appropriate dilation, provides reliable boundaries for the editing model to ensure accurate representation and minimizing errors. These examples underscore \emph{X-Planner}'s ability to deliver precise and adaptive control signals for complex image editing tasks.

\begin{figure*}[!t]
    \centering
    \includegraphics[width=1\linewidth]{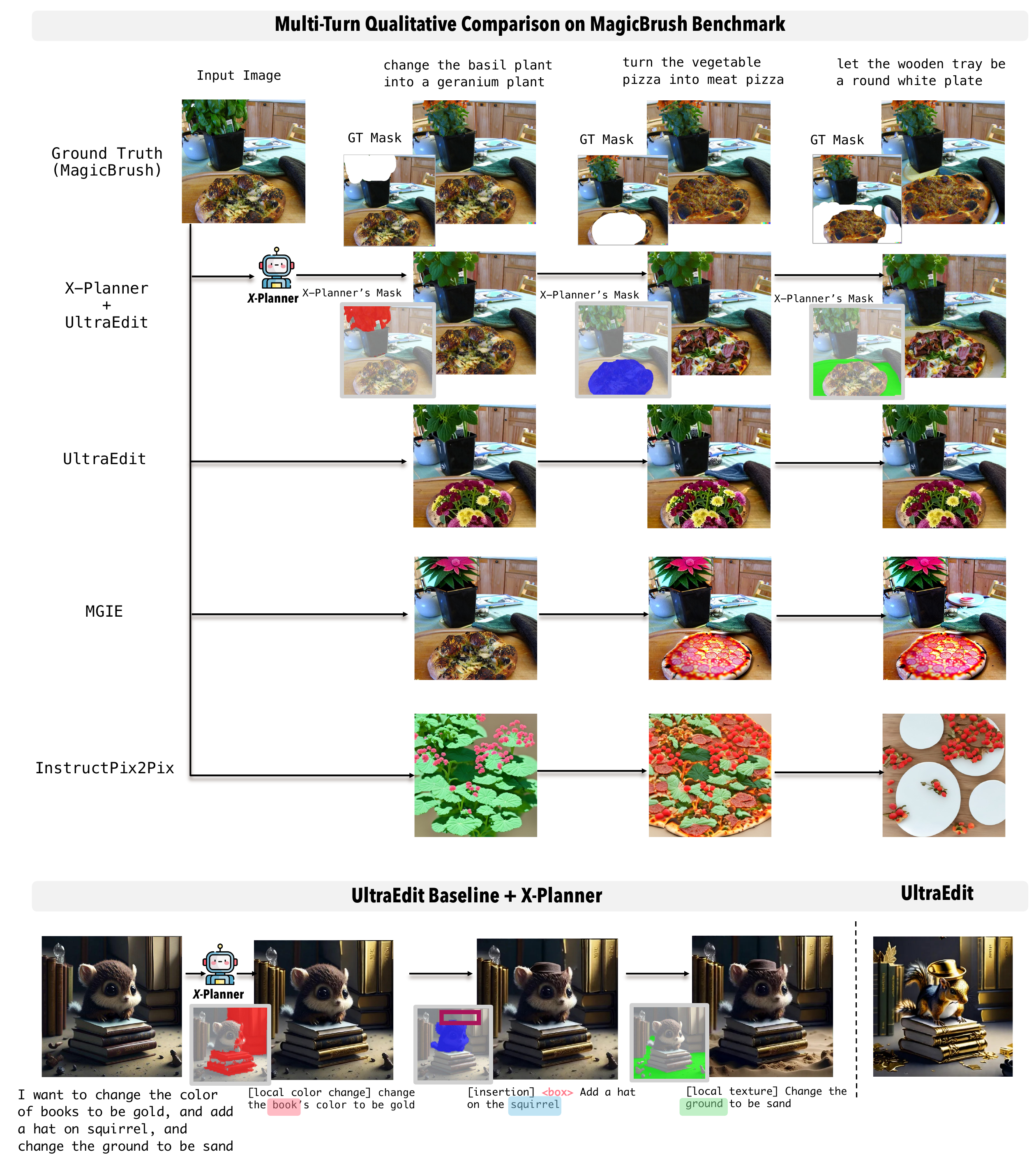}
    \vspace{-10pt}
    \caption{\textbf{Qualitative Comparison on MagicBrush, and on COMPIE Benchmark between \emph{X-Planner} and UltraEdit (I).} This figure presents: (1) a comparison of \emph{X-Planner} with several baseline methods for multi-turn editing on the MagicBrush dataset and (2) a qualitative comparison with the UltraEdit baseline on the COMPIE benchmark. For multi-turn editing, \emph{X-Planner}, guided by its generated masks, demonstrates superior identity preservation. In contrast, many baseline methods, especially InstructPix2Pix, tend to overedit the image due to a lack of control.}

    \label{fig:supp-qual-ue}
\end{figure*}

In Figure~\ref{fig:supp-qual-ue}, we show (1) a comparative evaluation of \emph{X-Planner} against several baseline methods for multi-turn editing on the MagicBrush dataset and (2) a qualitative comparison with the UltraEdit baseline on the COMPIE benchmark. In the context of multi-turn editing, \emph{X-Planner} stands out by leveraging its generated masks to achieve superior identity preservation. In contrast, many baseline methods, particularly InstructPix2Pix, often over-edit the image, highlighting their lack of fine-grained control and precision. These results underscore \emph{X-Planner}’s ability to maintain consistency and accuracy across iterative edits.

\begin{figure*}[!t]
    \centering
    \includegraphics[width=1\linewidth]{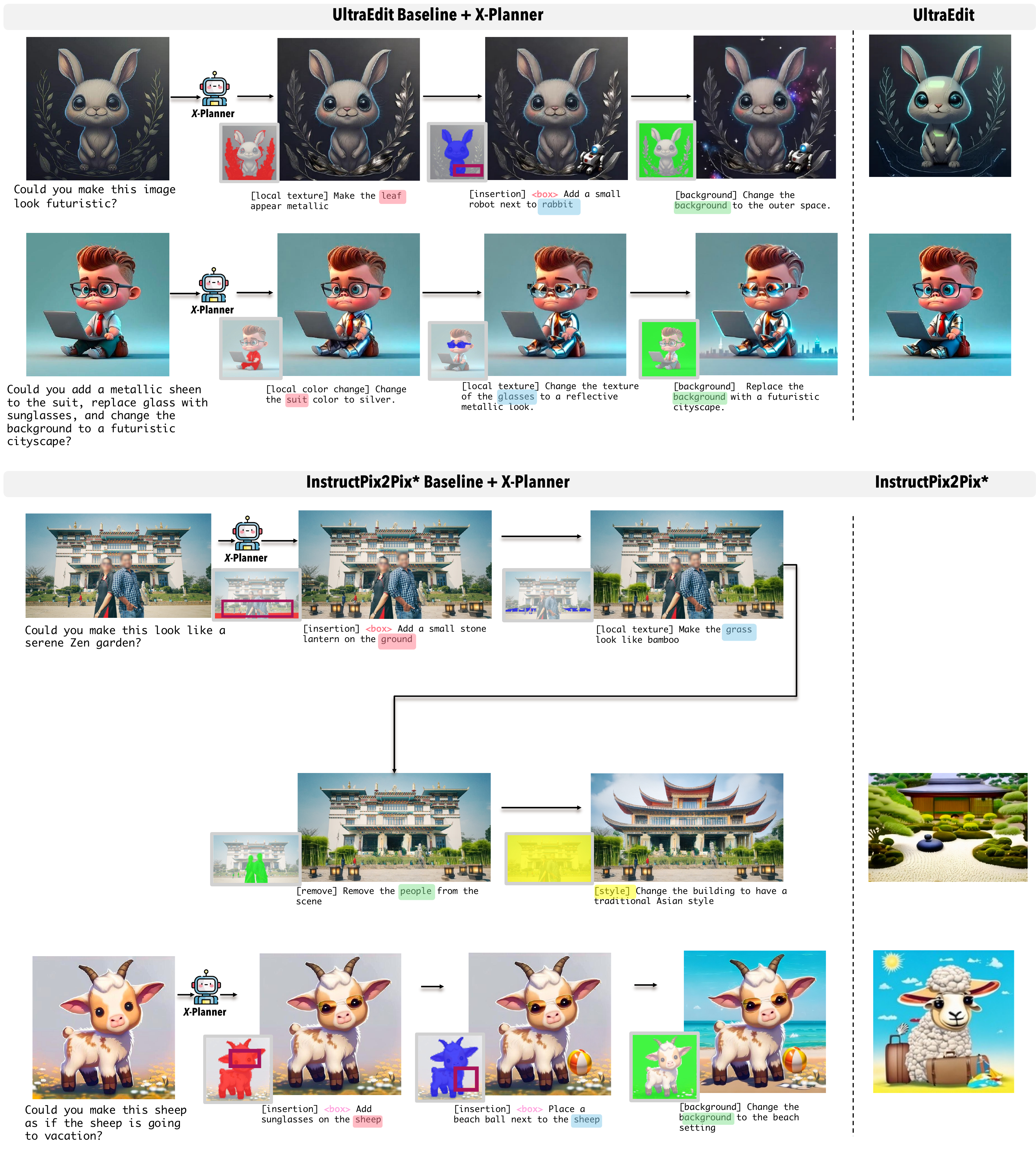}
    \caption{\textbf{Qualitative Comparison between \emph{X-Planner} and both InstructPix2Pix* (I) and UltraEdit Baseline (II).} This figure compares results from both InstructPix2Pix* and UltraEdit Baseline with and without the integration of \emph{X-Planner}. These examples highlight the advantages of \emph{X-Planner}'s instruction decomposition and mask controllability over the baseline which is directly given with complex instruction.}

    \label{fig:supp-qual-ie-part1}
\end{figure*}

\begin{figure*}[!t]
    \centering
    \includegraphics[width=1\linewidth]{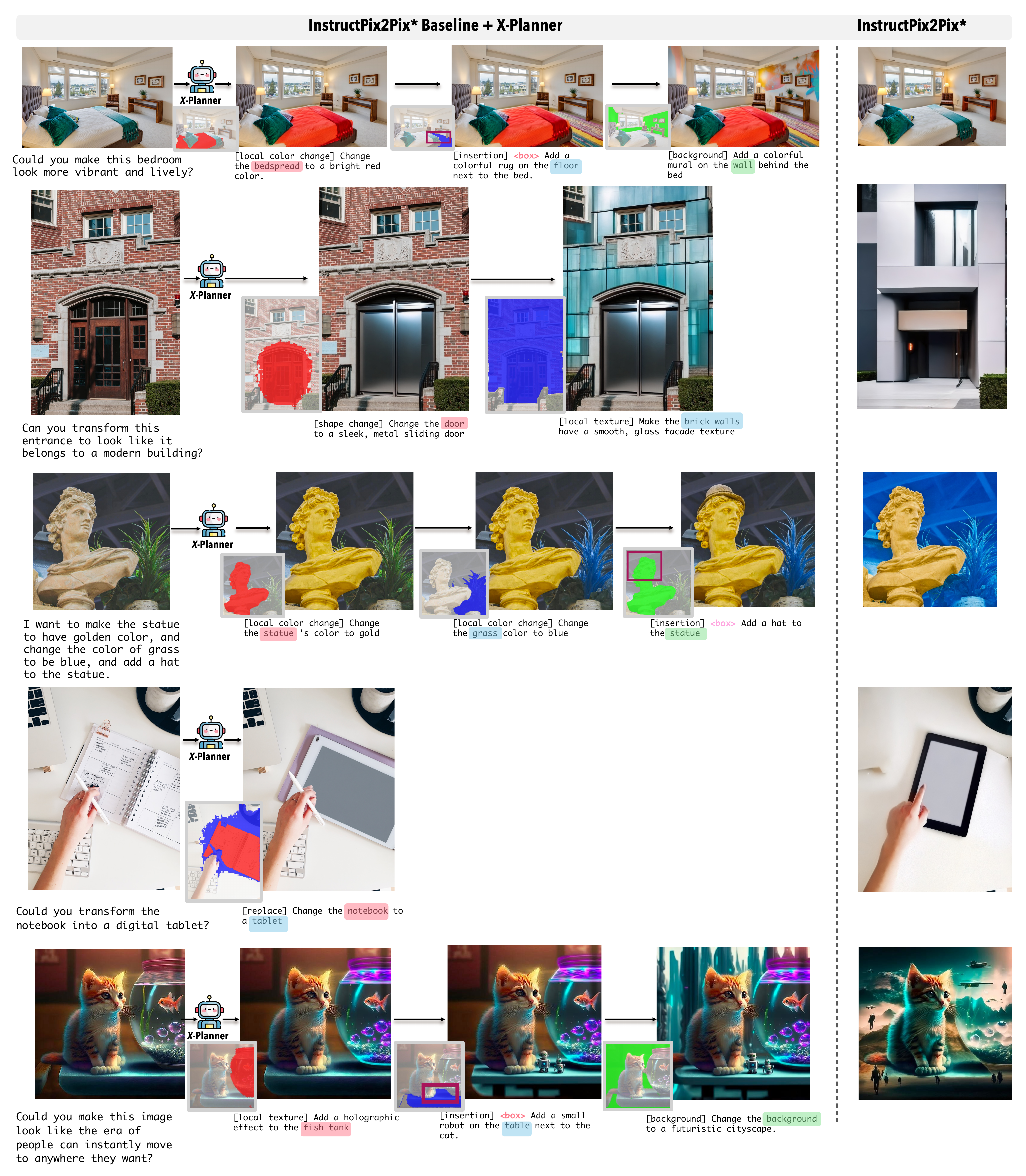}
    \caption{\textbf{Qualitative Comparison between \emph{X-Planner} and InstructPix2Pix* Baseline (II).} This figure compares results from InstructPix2Pix* with and without the integration of \emph{X-Planner}. For the [replace] edit type in Rows 4, \emph{X-Planner} generates a dilated segmentation mask for the replaced region to better accommodate the editing model. Similarly, in Row 2, the [shape change] edit includes a dilated mask to account for potential changes. These examples highlight the advantages of \emph{X-Planner}'s instruction decomposition and mask controllability over the baseline which is directly given with complex instruction.}

    \label{fig:supp-qual-ie-part2}
\end{figure*}

In Figure~\ref{fig:supp-qual-ie-part1} and Figure~\ref{fig:supp-qual-ie-part2}, we present a comparison between InstructPix2Pix* results with and without the integration of \emph{X-Planner}. For the [replace] edit type, as shown in Rows 4 in Figure~\ref{fig:supp-qual-ie-part2}, \emph{X-Planner} produces a carefully dilated segmentation mask for the replaced region, enabling the editing model to execute the changes more effectively. Likewise, in Row 2, the [shape change] edit incorporates a dilated mask, accommodating potential alterations in the object's shape and ensuring precise adjustments. These examples demonstrate the distinct advantages of \emph{X-Planner}’s instruction decomposition and mask generation capabilities, providing enhanced control and accuracy compared to the baseline, which relies on directly processing complex instructions without decomposition.

\section{Non-Rigid Edits on \emph{X-Planner}}
\label{sec:supp-non-rigid}

In Figure~\ref{fig:supp-non-rigid-edits}, we present a diverse set of \textbf{non-rigid and compositional edits} using three different editing models: InstructPix2Pix*, GPT-4o~\cite{achiam2023gpt}, and IC-Edit~\cite{zhang2025context}. These examples demonstrate the plug-and-play flexibility of our \emph{X-Planner}, which can integrate with existing editing models and enable them to perform complex edits—without requiring training on such complex instructions—by decomposing them into simpler, model-friendly steps.

However, as our framework relies on external editing models for actual image generation, the final results are sometimes bounded by their limitations. For example, GPT-4o often fails to preserve fine identity details across steps, particularly in face edits (\eg, Row 4 in Figure ~\ref{fig:supp-non-rigid-edits}, see red boxes). Similarly, UltraEdit~\cite{zhao2024ultraedit} occasionally generates inserted objects that slightly exceed their designated bounding box regions (see Main Paper, Figure~\ref{fig:xplanner-compare}), especially in cat in AI world scenes. These limitations are inherent to the editing models instead of the planning framework.

\begin{table*}[h!]
    \caption{\textbf{Bounding Box Localization on MULAN Benchmark.} 
    \emph{X-Planner} achieves strong localization gains by combining masks with MLLM-predicted boxes. 
    Pseudo-labeling improves AP$_{50}$ by over 2$\times$ at K=1, and box enlargement further boosts recall at K=5. Best performance is achieved using both mask and box cues.}

    \centering
    \resizebox{\columnwidth}{!}{
    \begin{tabular}{l|l|cc|cc|cc}
    \toprule
    & \multicolumn{7}{c}{\textbf{Insertion Edit Task (n = 416)}}  \\
    \textbf{Method} & \textbf{Setting} & \multicolumn{2}{c}{K=1} & \multicolumn{2}{c}{K=3} & \multicolumn{2}{c}{K=5} \\
    \cmidrule(lr){3-4} \cmidrule(lr){5-6} \cmidrule(lr){7-8}
    & & IoU$\uparrow$ & AP$_{50}$$\uparrow$ & IoU$\uparrow$ & AP$_{50}$$\uparrow$ & IoU$\uparrow$ & AP$_{50}$$\uparrow$ \\
    \midrule

    \multirow{3}{*}{\emph{X-Planner}} & Mask Only & \cellcolor{blue!5}0.37 & \cellcolor{blue!5}0.34 & \cellcolor{blue!5}0.46 & \cellcolor{blue!5}0.37 & \cellcolor{blue!5}0.54 & \cellcolor{blue!5}0.38 \\
    
    & +Pseudo-Label & \cellcolor{blue!5}0.63 & \cellcolor{blue!5}0.70 & \cellcolor{blue!5}0.71 & \cellcolor{blue!5}0.69 & \cellcolor{blue!5}0.75 & \cellcolor{blue!5}0.69 \\
    
    & + Box Enlarge & \cellcolor{blue!5}\textbf{0.75} & \cellcolor{blue!5}0.77 & \cellcolor{blue!5}\underline{0.81} & \cellcolor{blue!5}0.82 & \cellcolor{blue!5}\textbf{0.86} & \cellcolor{blue!5}0.84 \\
    & + Mask \& Box & \cellcolor{blue!5}0.73 & \cellcolor{blue!5}\textbf{0.78} & \cellcolor{blue!5}\textbf{0.81} & \cellcolor{blue!5}\textbf{0.86} & \cellcolor{blue!5}0.85 & \cellcolor{blue!5}\textbf{0.86} \\
    \bottomrule
    \end{tabular}
    }
    \label{tab:supp-x-planner-box}
\end{table*}

\begin{table*}[!h]
    
    \centering

    \caption{\textbf{Quantitative Comparison on the Emu Edit Test.} \emph{X-Planner} significantly enhances UltraEdit’s editing quality by decomposing complex instructions and supplying additional control inputs. We report metrics including L1 distance (lower is better), \textit{CLIP$_{\text{im}}$} and \textit{CLIP$_{\text{out}}$} similarity (higher is better), and \textit{DINO} feature similarity. Compared to baseline methods such as InstructPix2Pix, MagicBrush, EmuEdit, and UltraEdit, \emph{X-Planner} variants achieve consistently better performance across all metrics. Notably, the best results are obtained when combining \emph{X-Planner}'s mask and bounding box with a diverse bag of editing models.}

    \resizebox{\textwidth}{!}{
    \begin{tabular}{lccccc}
    \toprule
     \textbf{Methods} & \textbf{Guidance Control Input} & L1$\downarrow$ & CLIP$_{im}$ $\uparrow$ & CLIP$_{out}$$\uparrow$ & DINO$\uparrow$ \\ 
    \midrule
                  InstructPix2Pix(450K) & No & 0.1213 & 0.8518 & 0.2742 & 0.7656 \\  
                  MagicBrush(450+20K) & No & 0.0652 & \textcolor{red}{0.9179} & 0.2763 & \textcolor{red}{0.8964} \\  
                  EmuEdit(10M) & No & 0.0895 & 0.8622 & \textbf{0.2843} & 0.8358 \\
                  OmniGen & No & - & 0.8360 & 0.2330 & 0.8040 \\
                  \midrule
                  \cellcolor{gray!30}UltraEdit (1M w/o region data) & \cellcolor{gray!30}No & \cellcolor{gray!30}0.0515 & \cellcolor{gray!30}0.8915 & \cellcolor{gray!30}0.2804 & \cellcolor{gray!30}0.8656 \\
                  
                  \cellcolor{gray!30}UltraEdit (3M w/o region data) & \cellcolor{gray!30}No & \cellcolor{gray!30}0.0713 & \cellcolor{gray!30}0.8446 & \cellcolor{gray!30}0.2832 & \cellcolor{gray!30}0.7937 \\

                  \midrule
                  
                  \cellcolor{gray!30}UltraEdit  & \cellcolor{gray!30}No & \cellcolor{gray!30}0.0611 & \cellcolor{gray!30}0.8627 & \cellcolor{gray!30}0.2802 & \cellcolor{gray!30}0.8079 \\

                  \cellcolor{blue!5}\emph{X-Planner} + UltraEdit  & \cellcolor{blue!5}\emph{X-Planner}'s Mask & \cellcolor{blue!5}0.0462 & \cellcolor{blue!5}0.9007 & \cellcolor{blue!5}0.2782 & \cellcolor{blue!5}0.8723 \\
                  \cellcolor{blue!5}\emph{X-Planner} + UltraEdit  & \cellcolor{blue!5}\emph{X-Planner}'s Seg. Mask + \emph{X-Planner}'s Bounding Box & \cellcolor{blue!5}0.0457 & \cellcolor{blue!5}0.9029 & \cellcolor{blue!5}0.2798 & \cellcolor{blue!5}\textbf{0.8766} \\ 
                  \cellcolor{blue!5}\emph{X-Planner} + Bag of Models & \cellcolor{blue!5}\emph{X-Planner}'s Seg. Mask + \emph{X-Planner}'s Bounding Box & \cellcolor{blue!5}\textbf{0.0443} & \cellcolor{blue!5}\textbf{0.9046} & \cellcolor{blue!5}0.2822 & \cellcolor{blue!5}0.8754 \\

    \bottomrule
    \end{tabular}
    }
\label{tab:emu_result}
\end{table*}

\begin{table*}[!h]
    \caption{\textbf{Effectiveness of MLLM-Based Verification and Correction on COMPIE Evaluation.} We evaluate \emph{X-Planner} with closed-loop verification using GPT-4o and InternVL2.5-38B as step-wise verifiers. Our method achieves consistent gains over baselines by correcting intermediate errors through re-generation. We report improvements in both instruction-image alignment ($MLLM_{ti}$) and visual consistency ($MLLM_{im}$), highlighting the benefits of MLLM-guided correction for complex multi-step editing.}
   
    \centering
    
    \resizebox{\textwidth}{!}{
    \begin{tabular}{lccccc|cc}
    \toprule
     \textbf{Methods} & \textbf{Guidance \& Error Verify} & L1$\downarrow$ & CLIP$_{im}$ $\uparrow$ & CLIP$_{out}$$\uparrow$ & DINO$\uparrow$ & MLLM$_{ti}$ $\uparrow$ & MLLM$_{im}$ $\uparrow$ \\ 
    \midrule
    SmartEdit & No & 0.2764 & 0.7713 & 0.2512 & 0.6044 & 0.6511 & 0.5347 \\  
      MGIE & No & 0.2988 & 0.7692 & 0.2498 & 0.5981 & 0.6408 & 0.5288 \\  
      \midrule
      \cellcolor{gray!30}UltraEdit (UE) & \cellcolor{gray!30}No & \cellcolor{gray!30}0.1292 & \cellcolor{gray!30}0.7688 & \cellcolor{gray!30}\textbf{0.2698} & \cellcolor{gray!30}0.6387 & \cellcolor{gray!30}0.6652 & \cellcolor{gray!30}0.5523 \\

      \cellcolor{blue!5}\emph{X-Planner}+UE & \cellcolor{blue!5}Decomp.+No Verification & \cellcolor{blue!5}0.1253 & \cellcolor{blue!5}0.7767 & \cellcolor{blue!5}0.2621 & \cellcolor{blue!5}0.6435 & \cellcolor{blue!5}0.6894 & \cellcolor{blue!5}0.5593 \\ 
      
      \cellcolor{blue!5}\emph{X-Planner}+UE & \cellcolor{blue!5}Mask + Decomp.+No Verification & \cellcolor{blue!5}0.1188 & \cellcolor{blue!5}0.7875 & \cellcolor{blue!5}0.2569 & \cellcolor{blue!5}0.6599 & \cellcolor{blue!5}0.7061 & \cellcolor{blue!5}0.5744 \\ 
      
      \cellcolor{blue!10}\emph{X-Planner}+UE (GPT-4o) & \cellcolor{blue!10}Mask+Decomp.+Verification (max: 1) & \cellcolor{blue!10}0.1175 & \cellcolor{blue!10}0.7853 & \cellcolor{blue!10}0.2563 & \cellcolor{blue!10}0.6612 & \cellcolor{blue!10}0.7113 & \cellcolor{blue!10}0.5798  \\

        \cellcolor{blue!10}\emph{X-Planner}+UE (GPT-4o) & \cellcolor{blue!10}Mask+Decomp.+Verification (max: 4) & \cellcolor{blue!10}0.1163 & \cellcolor{blue!10}\textbf{0.7942} & \cellcolor{blue!10}0.2574 & \cellcolor{blue!10}\textbf{0.6673} & \cellcolor{blue!10}\textbf{0.7308} & \cellcolor{blue!10}0.5936 \\
        
        \cellcolor{blue!10}\emph{X-Planner}+UE(InternVL) & \cellcolor{blue!10}Mask+Decomp.+Verification (max: 1) & \cellcolor{blue!10}0.1180 & \cellcolor{blue!10}0.7861 & \cellcolor{blue!10}0.2559 & \cellcolor{blue!10}0.6632 & \cellcolor{blue!10}0.7128 & \cellcolor{blue!10}0.5822 \\
        
        \cellcolor{blue!10}\emph{X-Planner}+UE(InternVL) & \cellcolor{blue!10}Mask+Decomp.+Verification (max: 4) & \cellcolor{blue!10}\textbf{0.1160} & \cellcolor{blue!10}0.7901 & \cellcolor{blue!10}0.2571 & \cellcolor{blue!10}0.6647 & \cellcolor{blue!10}0.7258 & \cellcolor{blue!10}\textbf{0.5955} \\

    \bottomrule
    \end{tabular}
    }
\label{tab:verify_result}
\end{table*}

\begin{table*}[t]
    \centering
    \caption{\textbf{Quantitative Comparison on the MagicBrush Test Set (\emph{X-Planner} Trained from GPT-4o vs Pixtral-Large Generated Data).} We show single-turn (left) and multi-turn (right) performance. Our approach, using predicted masks and boxes, rivals or outperforms UltraEdit with human labels. "Bag of Models" uses PowerPaint~\cite{shen2024empowering} for removal, InstructDiff~\cite{geng2024instructdiffusion} for style, and UltraEdit otherwise.}
    \vspace{4pt}
    
    \begin{subtable}[t]{0.48\textwidth}
    \centering
    \caption{Single-Turn Editing}
    \resizebox{\textwidth}{!}{
    \begin{tabular}{lccccc}
    \toprule
    \textbf{Methods} & \textbf{Control} & L1$\downarrow$ & L2$\downarrow$ & CLIP-I$\uparrow$ & DINO$\uparrow$ \\
    \midrule
    \cellcolor{gray!30}UltraEdit (UE) & \cellcolor{gray!30}No   & \cellcolor{gray!30}0.0614 & \cellcolor{gray!30}0.0181 & \cellcolor{gray!30}0.9197 & \cellcolor{gray!30}0.8804 \\
    \cellcolor{gray!30}UltraEdit (UE) & \cellcolor{gray!30}Human Mask & \cellcolor{gray!30}0.0575 & \cellcolor{gray!30}0.0172 & \cellcolor{gray!30}0.9307 & \cellcolor{gray!30}0.8982 \\
    \midrule
    \multicolumn{6}{c}{\textbf{\emph{X-Planner} (GPT-4o Generated Training Data)}} \\
    \cellcolor{blue!5}\emph{X-Planner}+UE & \cellcolor{blue!5}Mask & \cellcolor{blue!5}0.0528 & \cellcolor{blue!5}0.0171 & \cellcolor{blue!5}0.9281 & 0.8900 \\
    \cellcolor{blue!5}\emph{X-Planner}+UE & \cellcolor{blue!5}Mask+Box & \cellcolor{blue!5}0.0513 & \cellcolor{blue!5}0.0168 & \cellcolor{blue!5}0.9312 & \cellcolor{blue!5}0.8959 \\
    \cellcolor{blue!5}\emph{X-Planner}+Bag of Models & \cellcolor{blue!5}Mask+Box & \cellcolor{blue!5}0.0511 & \cellcolor{blue!5}0.0172 & \cellcolor{blue!5}0.9331 & \cellcolor{blue!5}0.8970 \\
    \midrule
    \multicolumn{6}{c}{\textbf{\emph{X-Planner} (Pixtral Generated Training Data)}} \\
    \cellcolor{blue!5}\emph{X-Planner}+UE & \cellcolor{blue!5}Mask & \cellcolor{blue!5}0.0529 & \cellcolor{blue!5}0.0173 & \cellcolor{blue!5}0.9300 & \cellcolor{blue!5}0.8908 \\
    \cellcolor{blue!5}\emph{X-Planner}+UE & \cellcolor{blue!5}Mask+Box & \cellcolor{blue!5}0.0508 & \cellcolor{blue!5}\textbf{0.0165} & \cellcolor{blue!5}0.9324 & \cellcolor{blue!5}0.8983 \\
    \cellcolor{blue!5}\emph{X-Planner}+Bag of Models & \cellcolor{blue!5}Mask+Box & \cellcolor{blue!5}\textbf{0.0509} & \cellcolor{blue!5}0.0174 & \cellcolor{blue!5}\textbf{0.9342} & \cellcolor{blue!5}\textbf{0.8985} \\
    \bottomrule
    \end{tabular}
    }
    \end{subtable}
    \hfill
    \begin{subtable}[t]{0.48\textwidth}
    \centering
    \caption{Multi-Turn Editing}
    \resizebox{\textwidth}{!}{
    \begin{tabular}{lccccc}
    \toprule
    \textbf{Methods} & \textbf{Control} & L1$\downarrow$ & L2$\downarrow$ & CLIP-I$\uparrow$ & DINO$\uparrow$ \\
    \midrule
    \cellcolor{gray!30}UE w/o region & \cellcolor{gray!30}No  & \cellcolor{gray!30}0.0780 & \cellcolor{gray!30}0.0246 & \cellcolor{gray!30}0.8954 & \cellcolor{gray!30}0.8322 \\
    \cellcolor{gray!30}UE w/ region  & \cellcolor{gray!30}Human Mask & \cellcolor{gray!30}0.0745 & \cellcolor{gray!30}0.0236 & \cellcolor{gray!30}0.9045 & \cellcolor{gray!30}0.8505 \\
    \midrule
    \multicolumn{6}{c}{\textbf{\emph{X-Planner} (GPT-4o Generated Training Data)}} \\
    \cellcolor{blue!5}\emph{X-Planner}+UE & \cellcolor{blue!5}Mask & \cellcolor{blue!5}0.0679 & \cellcolor{blue!5}0.0227 & \cellcolor{blue!5}0.9025 & \cellcolor{blue!5}0.8423 \\
    \cellcolor{blue!5}\emph{X-Planner}+UE & \cellcolor{blue!5}Mask+Box & \cellcolor{blue!5}0.0668 & \cellcolor{blue!5}0.0226 & \cellcolor{blue!5}0.9047 & \cellcolor{blue!5}0.8475 \\
    \cellcolor{blue!5}\emph{X-Planner}+Bag of Models & \cellcolor{blue!5}Mask+Box & \cellcolor{blue!5}0.0665 & \cellcolor{blue!5}0.0223 & \cellcolor{blue!5}0.9079 & \cellcolor{blue!5}0.8508 \\
    \midrule
    \multicolumn{6}{c}{\textbf{\emph{X-Planner} (Pixtral Generated Training Data)}} \\
    \cellcolor{blue!5}\emph{X-Planner}+UE & \cellcolor{blue!5}Mask & \cellcolor{blue!5}0.0685 & \cellcolor{blue!5}0.0230 & \cellcolor{blue!5}0.9031 & \cellcolor{blue!5}0.8429 \\
    \cellcolor{blue!5}\emph{X-Planner}+UE & \cellcolor{blue!5}Mask+Box & \cellcolor{blue!5}0.0669 & \cellcolor{blue!5}0.0225 & \cellcolor{blue!5}0.9057 & \cellcolor{blue!5}0.8471 \\
    \cellcolor{blue!5}\emph{X-Planner}+Bag of Models & \cellcolor{blue!5}Mask+Box & \cellcolor{blue!5}\textbf{0.0661} & \cellcolor{blue!5}\textbf{0.0222} & \cellcolor{blue!5}\textbf{0.9083} & \cellcolor{blue!5}\textbf{0.8514} \\
    \bottomrule
    \end{tabular}
    }
    \end{subtable}
    \label{tab:magicbrush_pixtral_result}
\end{table*}

\begin{table*}[!t]
    
    \centering
    \caption{\textbf{Quantitative Comparison on the COMPIE Benchmark (\emph{X-Planner} Trained from GPT4o vs Pixtral-Large Generated Data).} X-Planner significantly improves the editing performance of UltraEdit and InstructPix2Pix* by decomposing complex instructions and providing control guidance inputs (e.g., segmentation masks). To overcome the limitations of $CLIP_{out}$ in handling complex instructions, we utilize an MLLM as an alternative evaluation metric to highlight capabilities of \emph{X-Planner}.}
    
    \vspace{-5pt}

    \resizebox{\textwidth}{!}{
    \begin{tabular}{lccccc|cc}
    \toprule
     \textbf{Methods} & \textbf{Guidance Control Input} & L1$\downarrow$ & CLIP$_{im}$ $\uparrow$ & CLIP$_{out}$$\uparrow$ & DINO$\uparrow$ & MLLM$_{ti}$ $\uparrow$ & MLLM$_{im}$ $\uparrow$ \\ 
    \midrule
                SmartEdit & No & 0.2764 & 0.7713 & 0.2512 & 0.6044 & 0.6511 & 0.5347 \\  
                  MGIE & No & 0.2988 & 0.7692 & 0.2498 & 0.5981 & 0.6408 & 0.5288 \\  
                  \midrule
                  \cellcolor{gray!30}UltraEdit & \cellcolor{gray!30}No & \cellcolor{gray!30}0.1292 & \cellcolor{gray!30}0.7688 & \cellcolor{gray!30}\textbf{0.2698} & \cellcolor{gray!30}0.6387 & \cellcolor{gray!30}0.6652 & \cellcolor{gray!30}0.5523 \\

                  \multicolumn{8}{c}{\textbf{\emph{X-Planner} Trained from GPT4o Generated Data}} \\ 

                  \cellcolor{blue!5}\emph{X-Planner} + UltraEdit & \cellcolor{blue!5}\emph{X-Planner}'s Decomposed Instruction & \cellcolor{blue!5}0.1253 & \cellcolor{blue!5}0.7767 & \cellcolor{blue!5}0.2621 & \cellcolor{blue!5}0.6435 & \cellcolor{blue!5}0.6894 & \cellcolor{blue!5}0.5593 \\

                  \cellcolor{blue!5}\emph{X-Planner} + UltraEdit & \cellcolor{blue!5}\emph{X-Planner}'s Mask + Decomposed Instruction & \cellcolor{blue!5}\textbf{0.1188} & \cellcolor{blue!5}\textbf{0.7875} & \cellcolor{blue!5}0.2569 & \cellcolor{blue!5}\textbf{0.6599} & \cellcolor{blue!5}0.7061 & \cellcolor{blue!5}0.5744 \\ 

                  \midrule
                  \multicolumn{8}{c}{\textbf{\emph{X-Planner} Trained from Pixtral-Large Generated Data}} \\ 

                  \cellcolor{blue!5}\emph{X-Planner} + UltraEdit & \cellcolor{blue!5}\emph{X-Planner}'s Decomposed Instruction & \cellcolor{blue!5}0.1261 & \cellcolor{blue!5}0.7744 & \cellcolor{blue!5}0.2630 & \cellcolor{blue!5}0.6428 & \cellcolor{blue!5}0.6904 & \cellcolor{blue!5}0.5626 \\

                  \cellcolor{blue!5}\emph{X-Planner} + UltraEdit & \cellcolor{blue!5}\emph{X-Planner}'s Mask + Decomposed Instruction & \cellcolor{blue!5}0.1207 & \cellcolor{blue!5}0.7853 & \cellcolor{blue!5}0.2584 & \cellcolor{blue!5}0.6577 & \cellcolor{blue!5}\textbf{0.7102} & \cellcolor{blue!5}\textbf{0.5765} \\ 
                  
                  \midrule

                  \cellcolor{gray!30}InstructPix2Pix* & \cellcolor{gray!30}No & \cellcolor{gray!30}0.1517 & \cellcolor{gray!30}0.8020 & \cellcolor{gray!30}\textbf{0.2666} & \cellcolor{gray!30}0.6988 & \cellcolor{gray!30}0.6727 & \cellcolor{gray!30}0.6160 \\

                  \multicolumn{8}{c}{\textbf{\emph{X-Planner} Trained from GPT4o Generated Data}} \\ 

                  \cellcolor{blue!5}\emph{X-Planner} + InstructPix2Pix* & \cellcolor{blue!5}\emph{X-Planner}'s Decomposed Instruction & \cellcolor{blue!5}0.1458 & \cellcolor{blue!5}0.8143 & \cellcolor{blue!5}0.2641 & \cellcolor{blue!5}0.7114 & \cellcolor{blue!5}0.7072 & \cellcolor{blue!5}0.6277 \\

                  \cellcolor{blue!5}\emph{X-Planner} + InstructPix2Pix* & \cellcolor{blue!5}\emph{X-Planner}'s Mask + Decomposed Instruction & \cellcolor{blue!5}\textbf{0.1320} & \cellcolor{blue!5}0.8285 & \cellcolor{blue!5}0.2591 & \cellcolor{blue!5}0.7068 & \cellcolor{blue!5}0.7408 & \cellcolor{blue!5}0.6454 \\

                  \midrule
                   \multicolumn{8}{c}{\textbf{\emph{X-Planner} Trained from Pixtral-Large Generated Data}} \\

                  \cellcolor{blue!5}\emph{X-Planner} + InstructPix2Pix* & \cellcolor{blue!5}\emph{X-Planner}'s Decomposed Instruction & \cellcolor{blue!5}0.1460 & \cellcolor{blue!5}0.8141 & \cellcolor{blue!5}0.2655 & \cellcolor{blue!5}0.7122 & \cellcolor{blue!5}0.7088 & \cellcolor{blue!5}0.6295 \\

                  \cellcolor{blue!5}\emph{X-Planner} + InstructPix2Pix* & \cellcolor{blue!5}\emph{X-Planner}'s Mask + Decomposed Instruction & \cellcolor{blue!5}0.1325 & \cellcolor{blue!5}\textbf{0.8291} & \cellcolor{blue!5}0.2586 & \cellcolor{blue!5}\textbf{0.7077} & \cellcolor{blue!5}\textbf{0.7431} & \cellcolor{blue!5}\textbf{0.6488} \\
    
    \bottomrule
    \end{tabular}
    }
\vspace{10pt}
\label{tab:COMPIE_pixtral_result}
\end{table*}

\begin{figure*}[!t]
    \centering
    \includegraphics[width=0.9\linewidth]{figure/supp-non-rigid-edits.pdf}
    \caption{\textbf{Examples of Non-Rigid and Compositional Edits with Model-Specific Failure Cases.} We show diverse complex edits (especially the [shape change] edits) generated by plugging \emph{X-Planner} into three editing models: InstructPix2Pix*, GPT-4o~\cite{achiam2023gpt}, and IC-Edit~\cite{zhang2025context}. Despite not being trained on such complex instructions, these models can perform challenging edits thanks to our decomposition and localization planning. However, some failure cases arise from the editing models: GPT-4o struggles with identity preservation in face edits (e.g., Row 4, red boxes).}

    \label{fig:supp-non-rigid-edits}
\end{figure*}